\documentclass[11pt]{article}

\usepackage[preprint]{acl}

\usepackage{times}
\usepackage{latexsym}
\usepackage[T1]{fontenc}
\usepackage[utf8]{inputenc}
\usepackage{microtype}
\usepackage{inconsolata}

\usepackage{graphicx}
\usepackage{amsmath}
\usepackage{amsfonts}
\usepackage{booktabs}
\usepackage{xcolor}
\usepackage{url}
\usepackage{svg}

\usepackage{float}

\title{Linguistic Context Recodes Visual Representations in Vision-Language Models}

\author{Brian Song \\
  Brown University \\
  \texttt{brian\_song1@brown.edu} \\\And
  Michael A. Lepori \\
  Brown University \\
  \texttt{michael\_lepori@brown.edu} \\\And
  Ellie Pavlick \\
  Brown University \\
  \texttt{ellie\_pavlick@brown.edu}}

\begin{document}

\maketitle
\begin{abstract}
Goal-directed visual processing is a hallmark of human visual intelligence, resulting in representations that support downstream tasks such as categorization or search. Though vision-language models (VLMs) are often faced with these same tasks, their ability to recode visual representations when presented with goal-directed language remains poorly characterized. Indeed, prior work largely treats visual representations in VLMs as static repositories of visual information that are manipulated by language representations. In the present work, we provide evidence for two concrete instances of language-induced recoding of visual representations. First, we identify an abstract \textit{reference representation} that denotes which objects are goal-relevant under a natural language prompt. We extract contrastive steering vectors corresponding to this reference representation and demonstrate that they are causally implicated in model predictions. These reference representations are abstract in that they generalize to different objects, different task contexts, and even from synthetic to naturalistic images. Second, we demonstrate language-induced \textit{attribute modulation}: later layers selectively amplify goal-relevant attributes in visual representations of objects. We demonstrate this phenomenon across a range of different prompts. Finally, we provide a causal intervention that demonstrates that attribute modulation mediates a VLM's response distribution.  Together, our results support a more dynamic account of cross-modality processing in VLMs --- rather than vision tokens serving as static repositories of information, they are modulated to support queries articulated in language.

\end{abstract}

\section{Introduction}
Vision-language models (VLMs) are often framed as systems that ground language in visual representations: images inject visual information, and language queries retrieve or reason over it \citep{VQA, openai2024gpt4technicalreport, thrush2022winoground, campbell2024understanding}. 
 In keeping with this common understanding of VLMs, prior efforts to interpret cross-modal processing have often focused on understanding how language tokens read and manipulate information stored in vision representations \citep{assouel2026visual, golovanevsky2025pixels, hua2025vision}. Vision encoders --- including those that serve as backbones to contemporary VLMs --- have been shown to produce rich, structured representations that reflect factors such as shape, color, spatial relations, objects, and numerosity \citep{neo2025towards, lidoes, hasani2025counting, lepori2024beyond, cui2026dual}. This literature aims to understand how exactly these repositories of visual features are \textit{used} in the service of language generation. Furthermore, the asymmetry between text and vision is often reflected in the very architectures of VLMs: several VLMs require images to be prepended to text, reducing the ability of VLMs to integrate language context into visual processing \citep{liu2023visual,li2023blip}.

These prior results invite the question: are visual representations truly just static repositories of visual information, or are they instead dynamically recoded in response to language?
To address this, we study the impact of linguistic input on visual representations. Such top-down, goal-directed signals are  known to recode visual representations in biological vision systems, often by directing spatial or feature-based attention to particular aspects of a stimulus \citep{buschman2007top,freedman2001categorical,corbetta2002control}. Language can encode such goal-directed signals, and we hypothesize that VLMs recode vision representations in order to support goals that are specified in text.


\begin{figure*}
    \centering
    \includegraphics[width=.95\linewidth]{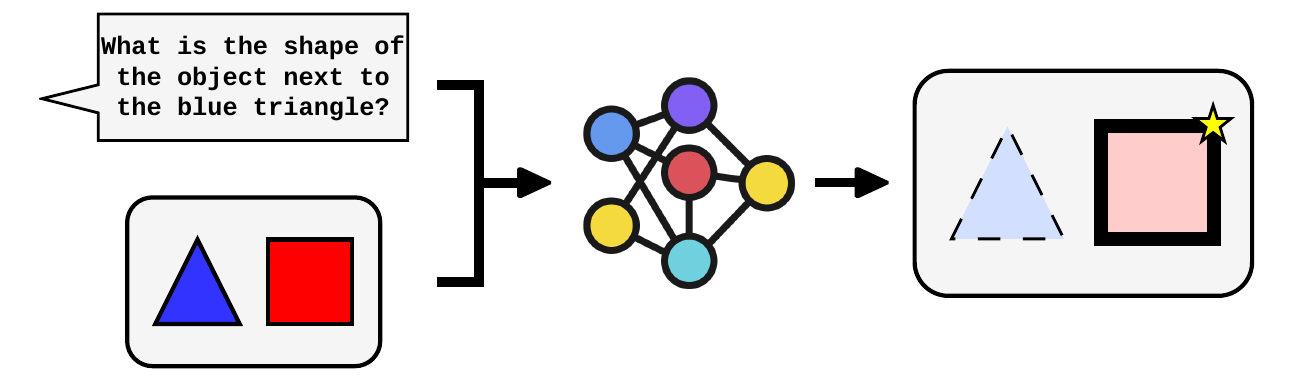}
    \caption{Goal-directed language input causes VLMs to recode visual representations. In this work, we hypothesize that text input will modify representations of images in support of goals (or queries) specified in language. In Section~\ref{sec:reference}, we find that VLMs add an abstract reference representation to visual representations of goal-relevant objects, visualized here as a star. In Section~\ref{sec:modulation}, we find that VLMs modulate representations of object features in a goal-directed manner, visualized here by increasing the thickness of the square outline, while decreasing the thickness of the triangle outline and  the saturation of the colors.}
    \label{fig:banner_fig}
\end{figure*}
We investigate two mechanisms by which linguistic input recodes visual representations, illustrated in Figure~\ref{fig:banner_fig}: (i) adding an abstract ``reference'' representation to visual representations corresponding to a goal-relevant object, and (ii) modulating existing features to emphasize those that are relevant to the current goal. Specifically, our contributions are:
\begin{enumerate}
\item We identify a linearly-decodable abstract ``reference'' representation within visual representations that dynamically marks whether the representation contains a goal-relevant object. We verify that reference representations reliably generalize across three different language-specified goals.
\item We find that language-specified goals result in visual representations whose features are modulated in service of these goals. 
\item In both instances, the recoding of visual representations has a causal influence on the VLMs' responses to the language-specified goals. Specifically, one can extract abstract reference representations and steer VLMs using them in multiple task contexts. Additionally, ablating goal-directed feature modulation reduces VLM confidence in their responses.
\end{enumerate}


\section{Related work}


Prior work has made progress on analyzing the internal mechanisms and 
representations implemented by vision-language models. These studies show that
VLMs do not treat images as unstructured inputs: visual patches can be linked to
specific linguistic outputs \citep{palit2023towards, golovanevsky-etal-2025-vlms},
visual and linguistic representations become partially aligned
\citep{wu2025the, nikankin2026same, dhimoila2026cross}, and object-level visual
representations encode information about entities and their attributes
\citep{lepori2024beyond, assouel2026visual}. Other work uses
probing and intervention-based analyses to study how VLMs resolve conflicts
between stored linguistic knowledge and visual evidence
\citep{golovanevsky2025pixels}, while recent work examines explicitly
conflicting multimodal inputs, including cases where image and text disagree and
models must determine which modality to rely on
\citep{zhu2024unraveling, deng2025wordsvisionvisionlanguagemodels, hua2025visionlanguagemodelsprocessconflicting}. Together, these
works suggest that VLMs contain rich visual representations that can be aligned
with, queried by, and sometimes overridden by language.

Most directly related to the present study, \citet{liu2025visual} demonstrates that text prefixes modify visual representation and lead to modest  performance improvements across three perceptual tasks, including segmentation of objects referred to by an image. This work thus suggests recoding of visual representations in the context of goal-directed language, but does not seek to characterize \textit{how} the visual representations changed.


\section{Methods}
\label{sec:methods}

\paragraph{Models}
We investigate Qwen 2.5 VL 7b \citep{bai2025qwen25vl} and InternVL3 8b \citep{zhu2025internvl3}, which are both transformer-based VLMs that enable text input to contextualize visual representations. Concretely, an input image is first encoded by a vision encoder and then projected into the language model's embedding space, yielding a sequence of image tokens that are concatenated with language tokens and processed jointly by the transformer's self attention layers. We report Qwen 2.5 VL results in the main section and InternVL3 results in Appendix~\ref{sec:internvl3_experiments}.

\paragraph{Synthetic Datasets}
We construct a synthetic image dataset to study how VLMs recode vision representations when provided with goal-directed language input. Each image contains several objects (three objects when studying reference representations, two or three when studying feature modulation), where each object is defined by a (color, shape) pair drawn from colors \{red, blue, green, yellow, purple, orange\} and shapes \{square, circle, triangle, heart, cross, star\}. No two objects in the same image have the same color or shape. Objects are aligned with a 4 $\times$ 4 grid of visual patches to enable object-level analysis.
\paragraph{Natural Image Dataset}
To test generalization of our reference representation analysis beyond simple synthetic settings, we construct a filtered subset of the COCO dataset \citep{lin2014microsoft}. Images are selected such that they contain at most four object instances with no repeated object classes. We additionally require that objects form a well-separated left-middle-right triplet in an image. From this filtered pool, we manually exclude images with cluttered or ambiguous scenes, yielding a natural image evaluation set of 100 images. Object segmentation masks provided by COCO are used to identify which vision tokens correspond to each object.

\section{Reference Representations}
\label{sec:reference}
Do VLMs recode visual representations of goal-relevant objects to denote that they are useful for answering a natural language query?  We find evidence of such \textit{reference representations} across several tasks through a series of linear probing and causal intervention analyses.

\subsection{Tasks}

Inspired by recent work characterizing abstract representations that bind tokens together in both language \citep{feng2024how} and vision models \citep{lidoes, tartaglini2026walk}, we seek to understand whether VLMs produce abstract representations that indicate the goal-relevance of a vision token. An abstract representation of reference would be transferable between distinct objects and distinct tasks. In this work, we consider three simple tasks using both synthetic and naturalistic data:

\begin{itemize}
\item \textbf{Counting:}
Queries the model for the number of objects that exhibit a given attribute, e.g., ``How many shapes are red in the image?''
Objects matching the queried attribute (e.g., red) are labeled as referents, while all others are non-referents.

\item \textbf{Yes/No:}
Queries the model for whether an image contains at least one object with a given attribute, e.g., ``Is there a circle shape in the image?''
Objects matching the queried attribute are labeled as referents, and all others are non-referents.

\item \textbf{Spatial:}
Queries the model for a description of an object based on its spatial relation (left/right/above/below) to another landmark object, e.g., ``What shapes are above the green square?''
Only the object in the specified spatial relation with the landmark is labeled as a referent. The shape that is neither the landmark nor the referent is labeled as a non-referent. Images are constructed so that exactly one object satisfies the relation.
\end{itemize}

\subsection{Reference Representations are Decodable}
\label{sec:referred_prompt}
Given an image, we extract and concatenate the vision tokens that contain each object. We train a binary classifier to predict whether that object is a referent or non-referent in the context of the language query. We train separate probes per layer and per task on a class-balanced dataset of 1000 images per task.

As a control, we attempt to train the same probe on vision representations of objects that were generated without prepending text that articulates the task. We experiment with three such ablations: an empty string, a prompt describing all objects in the image (without any query/goal), or a prompt describing three objects whose attributes are not present anywhere in the image. Prompt details are listed in Appendix~\ref{sec:probe_prompts}. We train separate linear probes for each prompt type to predict the original labels. 

We report results for the Spatial task  in Figure~\ref{fig:probes} and Counting and Yes/No in Appendix~\ref{sec:qwen_probe_extra}. For all tasks, in the presence of goal-direction language input, probe accuracy  starts near chance (50\%) in early layers and peaks in middle to late layers. Accuracy remains at chance across all layers for all controls, confirming that goal-directed language (rather than no language or merely mentioning an object) drives the representational change.

\begin{figure}[t]
  \centering
  \includegraphics[width=\columnwidth]{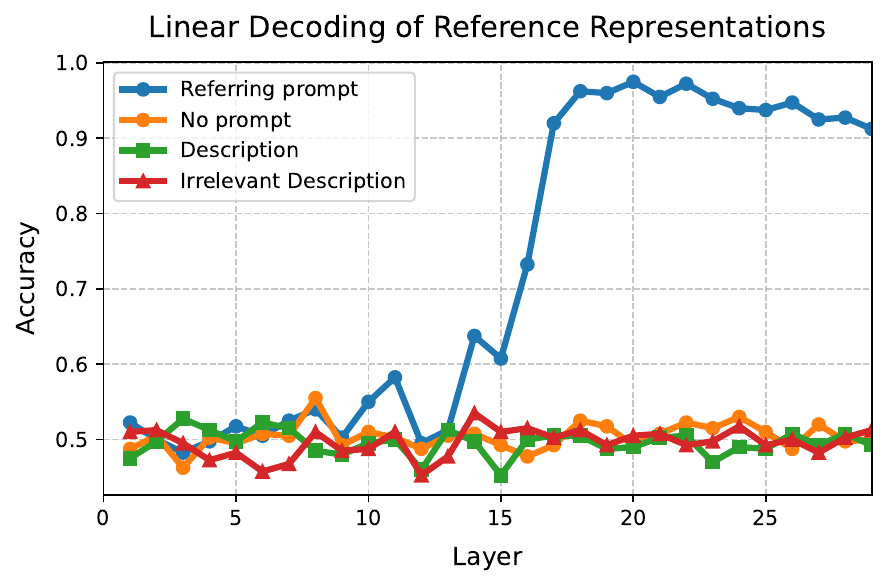}
  \caption{Linear probing accuracy for referent vs non-referent classification on the Spatial Task. Task-specific prompts yield accuracy that rises from chance in early layers and peaks in middle to late layers, while control prompts remain at chance throughout, confirming that reference from prompt drives recoding of visual representations.}
  \label{fig:probes}
\end{figure}

\subsection{Reference Representations are Causal}
High linear decoding accuracy suggests the existence of a reference representation, but does not indicate whether it is used by the VLM to answer the query. We assess the causal role of reference representations using contrastive steering vectors \citep{rimsky-etal-2024-steering}.

\paragraph{Constructing Steering Vectors}
We run two separate inference passes for each image, differing only in the prepended language prompt. Prompts differ only in which object is the referent. For each prompt, we extract hidden states corresponding to the objects in the image, and compute the difference between object representations when the underlying object is a referent vs. a non-referent. We average these difference vectors across multiple images to create a stable steering vector, and repeat this process for each layer and task. Steering vectors are unit normalized. 

\begin{figure}[t]
  \centering
  \includegraphics[width=\columnwidth]{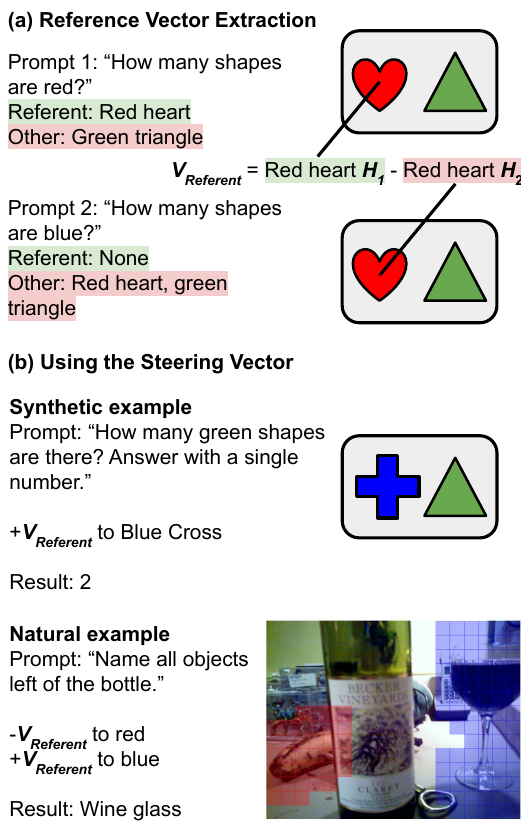}
  \caption{(a) Reference steering vector extraction for the spatial
    task. Given a contrastive prompt pair, the reference direction for
    each object is computed as the difference in vision token hidden
    states between the condition where it is a referent and where it is
    not, then averaged across examples. (b) Intervention examples on
    a synthetic image and a natural COCO image. For the latter, adding
    the reference vector to a non-referent object and subtracting it
    from the referent object flips the model's output to the alternate
    object.}
  \label{fig:steering}
\end{figure}

 Contrastive object pairs are constructed by swapping the queried attribute (for the counting and yes-no tasks) or spatial relation (for the spatial task). 
 See Figure~\ref{fig:steering}(a) for an illustration. We also included negation pairs (e.g., ``How many circle shapes?''\ vs.\ ``How many not circle shapes?'') to increase prompt diversity (See Appendix~\ref{sec:prompts} for the full list of prompt templates).




\begin{figure*}[t]
  \centering
  \includegraphics[width=\textwidth]{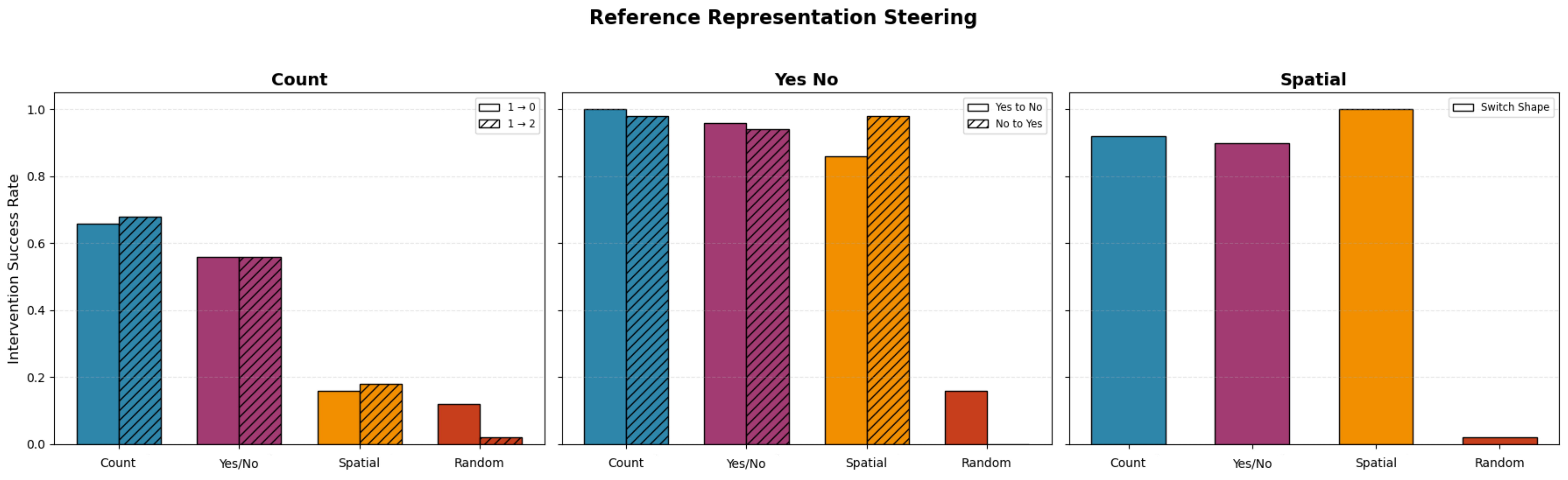}
  \caption{Synthetic dataset intervention success rates for task-specific steering vectors on Count, Yes/No, and Spatial tasks. Each bar shows the proportion of correct baseline predictions that are successfully flipped under intervention. Same-task vectors typically achieve the highest success rates, supporting the existence of a reference representation. Reference representations transfer across most tasks, indicating that these reference representations are abstract (i.e., not tied to a particular object or task). All interventions outperform a random vector baseline. }
  \label{fig:intervention_results}
\end{figure*}

\paragraph{Interventions}

Interventions are performed by adding or subtracting the (scaled) reference representation to the hidden states corresponding to the target object at layer $l$. Ideally, adding the reference representation would push the model towards considering the corresponding object as a goal-relevant object, and subtracting the reference representation would have the opposite effect. Example interventions are shown in Figure~\ref{fig:steering}(b) and full evaluation
prompts are listed in Appendix~\ref{sec:prompts}.  

Reference interventions have unique outcomes for each task. For the counting task, we intervene to transform the answer from 1 to 0, or from 1 to 2. For example, given the prompt ``How many green shapes are there?" and an image with a single green shape, we subtract the reference representation from the green shape or add the reference representation to a shape that is not green, respectively.

For the yes/no task, we intervene to flip the answer. Given the prompt "Is there a circle shape?" and an image where there is one circle, we subtract the reference representation from the circle shape. Given the same prompt coupled with an image that does not contain a circle, we add the reference vector.

For the spatial tasks, we intervene to flip the model's response from one shape to another. Consider a prompt that queries the model to describe ``the object to the left of the green circle'', and an image with a red circle to the left and a blue triangle to the right. We add the reference representation to the blue triangle and subtract it from the red circle.

Interventions are evaluated on 1000 examples for which the model's baseline prediction is correct. Baseline accuracy for Counting, Yes/No, and Spatial task is 94, 89, and 99\%, respectively. We sweep steering coefficients and apply interventions across different contiguous 15-layer blocks spanning mid-to-late model layers. We record the best hyperparameters on a validation set shown in Appendix~\ref{sec:hyperparameters} and report performance on a test set.


\paragraph{Intervention Results}
Results are shown in Fig.~\ref{fig:intervention_results}. Interventions reliably succeed, from a 60\% success rate for the counting task to ceiling success rates for the other tasks. As a baseline, we also steer using a norm-matched random vector. We find that these baseline vectors perform consistently worse than steering with reference representations.

\paragraph{Cross-Task Generalization}
Enriching visual representations to denote whether they are goal-relevant is a fundamental operation that is useful across distinct tasks. Do VLMs use generic reference representations to perform this operation, or are they task-specific? We investigate this by assessing whether reference representations derived from one task generalize to the others. We find that reference representations extracted from one task typically transfer effectively to other tasks, with the exception of spatial task reference representations transferring to the counting task. This suggests that VLMs construct abstract reference representations that can be applied in diverse contexts. 

\begin{figure}[t]
  \centering
  \includegraphics[width=\columnwidth]{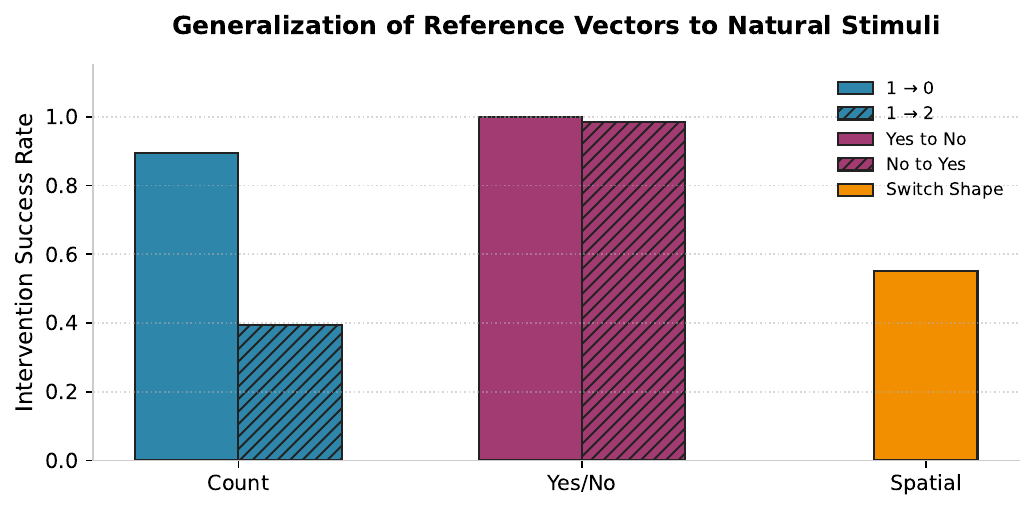}
  \caption{Naturalistic dataset intervention success rates using same-task reference vectors on Count, Yes/No, and Spatial tasks. Reference vectors generalize effectively from extremely simple, synthetic images to naturalistic images, suggesting that reference representations are a generic mechanism in VLMs. }
  \label{fig:coco_counting_results}
\end{figure}

\paragraph{Naturalistic Generalization}
We next assess whether the reference representations extracted from simple, synthetic stimuli generalize to more naturalistic stimuli. This would be further evidence that reference representations are a generic mechanism of visual recoding. In  Figure~\ref{fig:coco_counting_results}, we present results from intervening on naturalistic stimuli using reference representations extracted from synthetic stimuli corresponding to the same task. We include cross-task, cross-dataset generalization in Appendix~\ref{sec:cross task_intervention}.  Overall, the reference representations generalize effectively to natural images.



\section{Attribute modulation}
\label{sec:modulation}

The previous section demonstrated that goal-directed language endows visual representations of goal-relevant objects with a reference representation. We now ask whether language also changes how visual attributes are
encoded inside those representations. We call this effect
\textit{attribute modulation}. 

\subsection{Visual Concept Vectors}
\label{sec:concept_vectors}

To quantify how strongly a visual attribute is represented in the visual representations corresponding to that object, we construct visual concept vectors for each shape and color \citep{savietto2026geometryrepresentationalfailuresvision}. For a given attribute (a particular shape or particular color), we compute the mean visual representation of all objects that possess that attribute. We then subtract the mean visual representation over all objects, which isolates the direction that encodes the targeted attribute. Averages are computed from 1000 images without any text prefix, so the concept directions reflect vision-only
representations.

We measure the degree to which a set of representations corresponding to an object encodes a particular attribute by projecting each representation onto the corresponding concept vector. As a control, we also compute projections for \emph{absent} concepts --- shape or color
values that do not appear anywhere in the image. A formal definition of the concept vectors and projection computation is provided in Appendix~\ref{sec:appendix_concept_vectors}. For each experiment below we record mean projections across 5000 images.




\begin{figure}[t]
  \centering
  \includegraphics[width=\columnwidth]{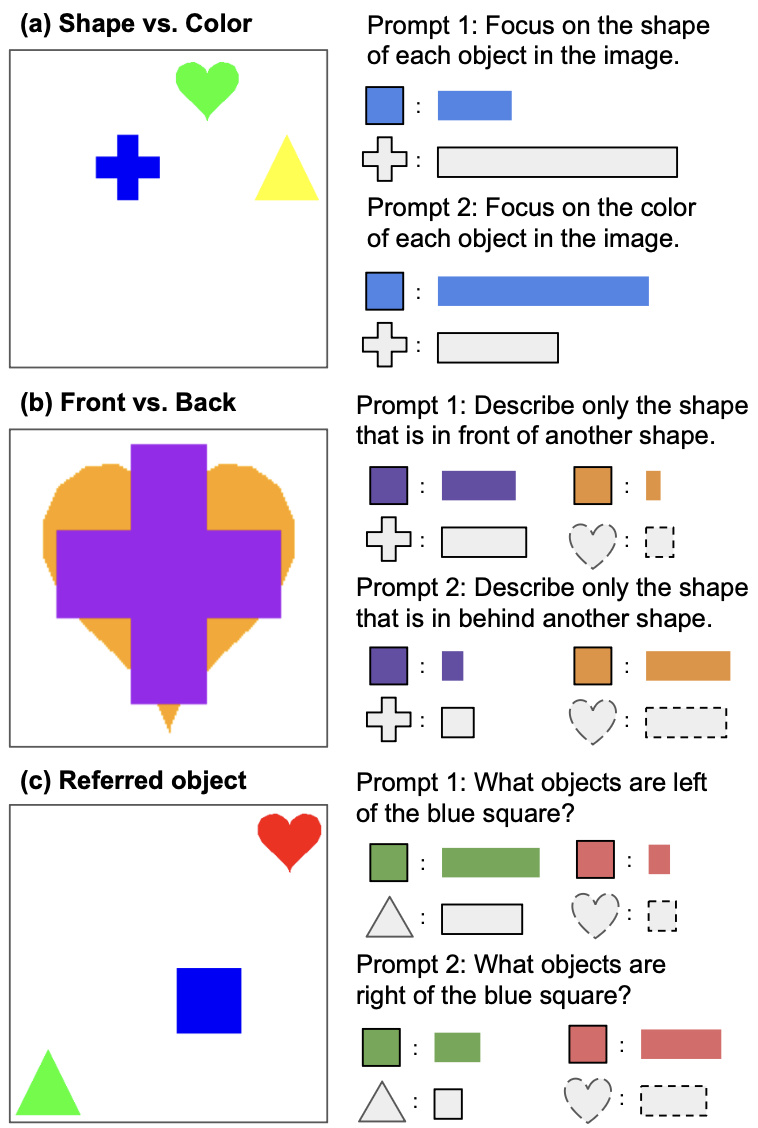}
  \caption{
  Overview of the attribute modulation experiments.
  \textbf{a)} Explicit shape vs.\ color modulation compares shape-focused and
  color-focused prompts by measuring projections onto each object's own shape and color.
  \textbf{b)} Front vs. back modulation uses an overlapping-object dataset to test
  whether referring to the front or back object changes projections onto that
  object's attributes, even though those attributes are not explicitly asked for.
  \textbf{c)} Referent object modulation uses spatial prompts to
  test whether the referent object's attributes are amplified in comparison to the non referent object's attributes.
}
\label{fig:priming_diagram}
\end{figure}

\subsection{Evidence for Attribute Modulation}
\paragraph{Explicit Shape and Color Modulation}
\label{sec:shape_color_priming}

We first test whether directly querying a specific attribute selectively amplifies
that attribute's representation. Using the synthetic dataset, we
compare each object's projections along the object's color and shape concept vectors under a shape-focused prompt (``Focus on the shape of each object in the
image.''), a color-focused prompt (``Focus on the color of each object in
the image.''), and the description prompt from Section~\ref{sec:referred_prompt} as a baseline (which lists the properties of all objects in the images). 

For each object, we measure how much
the shape and color-focused prompts change the object's attribute projections
relative to the baseline. We then visualize these changes in Figure~\ref{fig:shape_color_15_scatter} for a single layer, and include other layers in Appendix~\ref{sec:scatterplot}. Formal definitions of the modulation quantities are provided in Appendix~\ref{sec:appendix_shape_color_modulation}.
On average, the shape-focused prompt
increases projections onto the corresponding shape concept vector relative to the
baseline, while the color-focused prompt increases projections onto the corresponding
color concept vector. Furthermore, these shifts are attribute specific; a shape prompt predominantly increases projections on the shape vector \textit{over and above} the color vector, and vice versa. We provide extended analyses in Appendix~\ref{sec:rsa_appendix}.




\begin{figure}[t]
  \centering
  \includegraphics[width=\linewidth]{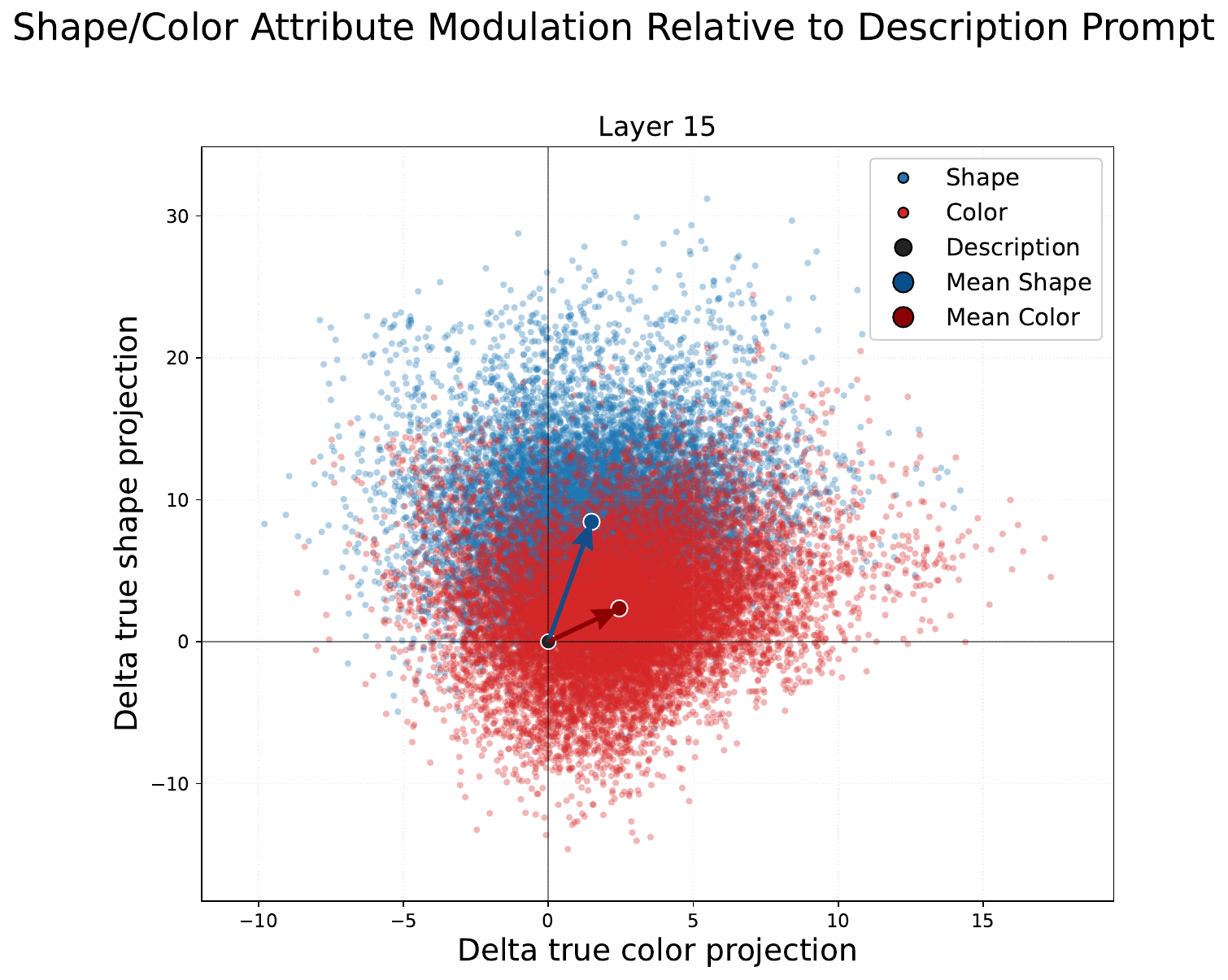}
  \caption{Shape/Color Attribute Modulation. We compare the projections along vectors corresponding to each object's shape/color attributes in the presence of a prompt which (1) emphasizes shape, (2) emphasizes color, or (3) generically describes objects in the image. We plot the difference in attribute projections for the same object in the presence of prompt (1) - (3) in blue, and plot the difference in attribute projections for the same object in the presence of prompt (2) - (3) in red. We find that prompts that emphasize shape increase projections along the corresponding shape vector, while minimally increasing projections along the color vector, and vice versa for prompts that emphasize color.}
\label{fig:shape_color_15_scatter}
\end{figure}

\paragraph{Front/Back Object Modulation}
\label{sec:front_back_priming}

\begin{figure}[t]
  \centering
  \includegraphics[width=\linewidth]{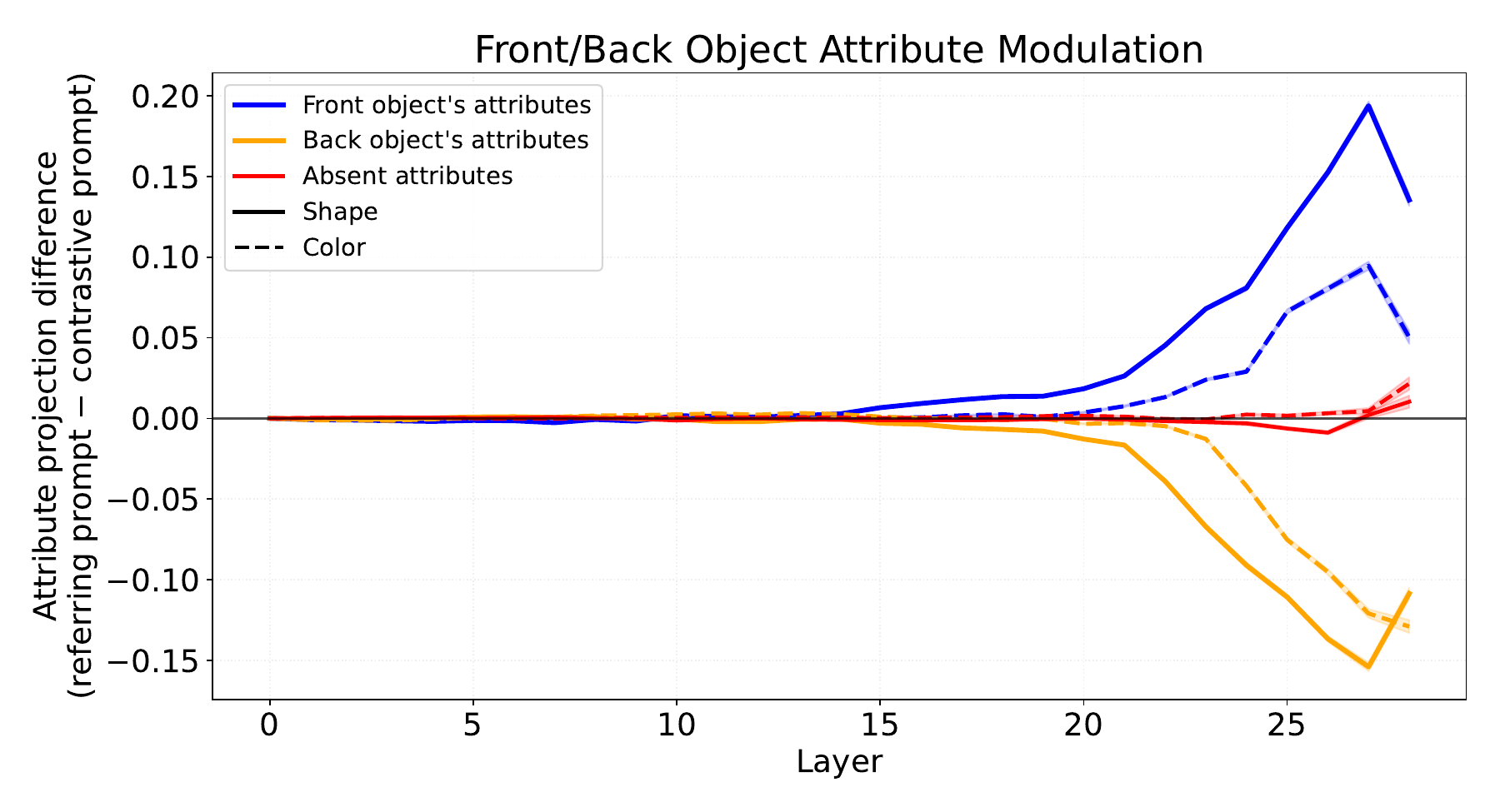}
  \includegraphics[width=\linewidth]{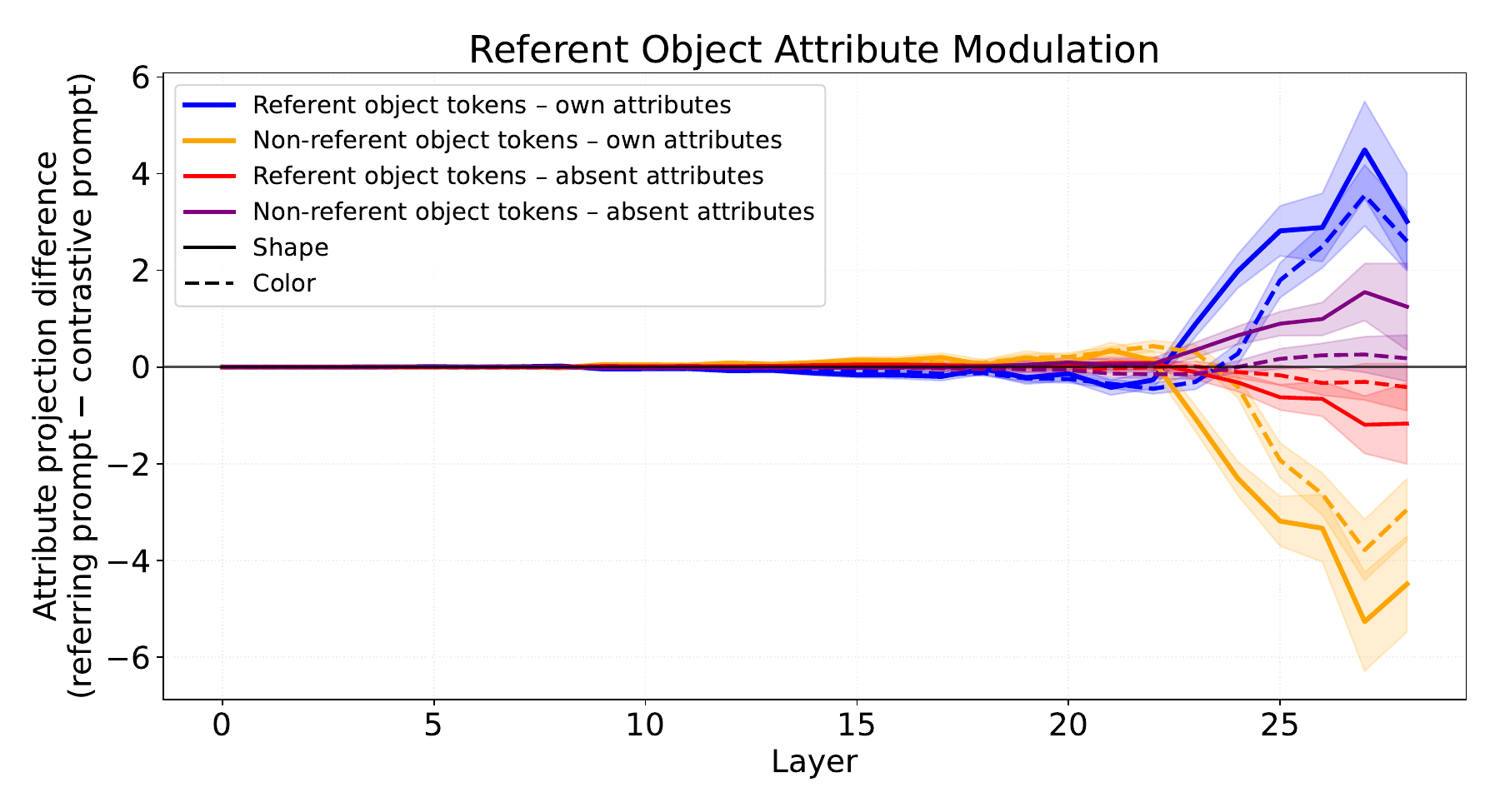}
  \caption{\textbf{Top:} Front/back attribute modulation using front/back prompt pairs on the overlapping object dataset. We visualize the difference in projections of all vision tokens onto concept vectors corresponding to the front object's attributes, the back object's attributes, and absent attribute controls when an image is coupled with (1) a front-focused prompt vs. (2) a back-focused prompt. Front-focused prompts increase projections onto the front object's attributes, while back focused prompts increase projections onto the back object's attributes. \textbf{Bottom:} Referent object modulation using spatial prompt pairs on the synthetic dataset. For each object, the referent and non referent, we project its visual tokens onto concept vectors corresponding to its own attributes and absent attribute controls. Referent object tokens exhibit increased projections onto their attributes, while projections onto absent attribute controls are suppressed. This  indicates that the VLM selectively sharpens the referent object's attribute representations.}
  \label{fig:spatial_depth_priming}
\end{figure}

We next test whether selecting an object can modulate its
attribute representations, even when those attributes are not explicitly queried.
We construct a new overlapping-object dataset (Figure~\ref{fig:priming_diagram}(b))
in which each image contains two overlapping objects --- one foreground and one
background --- with distinct shapes and colors.  Prompts select an object by
depth-ordering: ``Describe only the shape that is in front of another
shape.''\ versus ``Describe only the shape that is behind another shape.''
Because each object occupies nearly the entire image, we compute projections using all vision tokens. Specifically, we project these tokens onto the concept vectors corresponding to the front and back object's true shape and color. As a control, we also project these tokens onto concept vectors for colors and shapes that are not present in the image. We do this computation under both the front-focused and back-focused prompts, and measure the difference in their concept vector projections. Results are shown in Figure~\ref{fig:spatial_depth_priming} (Top). Front-focused prompts
increase projections onto foreground-object attributes, while back-focused
prompts increase projections onto background-object attributes. Absent concept controls
remain near zero, indicating that modulation remains tied to the selected object's attributes rather than reflecting a generic increase. Additional implementation details and the formal modulation definition are provided in Appendix~\ref{sec:appendix_front_back_modulation}.


\paragraph{Referent Object Modulation}
\label{sec:referred_priming}

We now test whether attribute modulation occurs when using a referring prompt pair from Section~\ref{sec:reference} by revisiting the spatial task. Given a fixed image, we generate a minimal prompt pair of the form (\texttt{What object is left of the blue square?}, \texttt{What object is right of the blue square?}). We arbitrarily choose one of these prompt to fix the referent vs. non-referent object labels. For each object (referent and non-referent), we project that object's vision tokens onto the concept vectors corresponding to 1) its own true attributes (shape and color) and 2) absent attribute controls. We measure the difference in concept vector projections between the two prompts, thus quantifying the effect of reference on attribute encoding. A formal definition of the referent and non-referent modulation quantities is provided in Appendix~\ref{sec:appendix_referent_modulation}.




Results are shown in Figure~\ref{fig:spatial_depth_priming} (Bottom). Starting in late layers,
the projections onto the referent object's shape and color become strongly positive, indicating amplification of its attributes. 
Meanwhile, projections onto attributes that are not present in the image remain comparatively small
throughout, showing that the effect is highly selective. Attribute modulation emerges several layers after the reference signal for the same spatial task
first becomes linearly decodable (See~Figure~\ref{fig:probes}), suggesting a two-stage
computation in which middle layers first identify the relevant object and later
layers selectively sharpen its attribute representations while suppressing
competing alternatives.

\subsection{Attribute Modulation is Causal}
\label{sec:freezing}

We have demonstrated evidence of attribute modulation in the later layers of VLMs, but does this visual recoding have any causal influence over a model's response to a query? In this section, we assess the causal role of attribute modulation.

\paragraph{Intervention}
To test whether late-layer attribute modulation causally contributes to answer confidence in a spatial task, we
freeze the referent object's vision token hidden states from layer $l$ onward to prevent further modulation. We measure the effect on the model's confidence in its response, both at the end of processing and over layers. We assess response confidence over layers using logit lens~\citep{nostalgebraist2020logit}, which projects the hidden state at layer $l$ to the vocabulary to produce token distributions at each layer. We apply logit lens at the final text token after appending ``Answer with just the object's color/shape.'' to elicit model responses. We compute the \textbf{logit gap} between the referent and non-referent attribute:
\begin{equation}
\mathrm{Gap}^{a}_{l}
=
M_l\big(a(o_{\mathrm{ref}})\big)
-
M_l\big(a(o_{\mathrm{nonref}})\big),
\end{equation}
where $M_l(a(o))$ is the logit of attribute $a$ of object $o$ at model layer $l$. Intuitively, this value indicates how confident a model is when choosing between the reference object's attribute vs. a different attribute that is also present in the image. We compare logit gaps when we freeze visual object representations (i.e., prevent attribute modulation) to the logit gap given by unmodified logits for that same (attribute, object, layer) tuple:
\begin{equation}
\Delta\mathrm{Gap}^{a}_{l}
=
\mathrm{Gap}^{a}_{l,\mathrm{freeze}}
-
\mathrm{Gap}^{a}_{l,\mathrm{baseline}}.
\end{equation}
Negative values indicate reduced confidence in the correct referent attribute. We run a complementary analysis that assesses the causal influence of modulation in Appendix~\ref{sec:concept_intervention}.

\paragraph{Results}
Logit gap results for the color attribute are shown in Figure~\ref{fig:freeze_color} while shape results are in Appendix~\ref{sec:freeze_shape_gap}. Freezing referent-object tokens reduces the final logit gap for both shape and color across all freeze points, confirming that continued processing at the referent object's position causally contributes to answer confidence. Critically, the effect persists even when freezing begins in the final layers of the network — the same layers where attribute modulation is most pronounced in Figure~\ref{fig:spatial_depth_priming}. This indicates that late-layer attribute amplification shapes the model's final answer. 

\begin{figure}[t]
    \centering
    \includegraphics[width=\columnwidth]{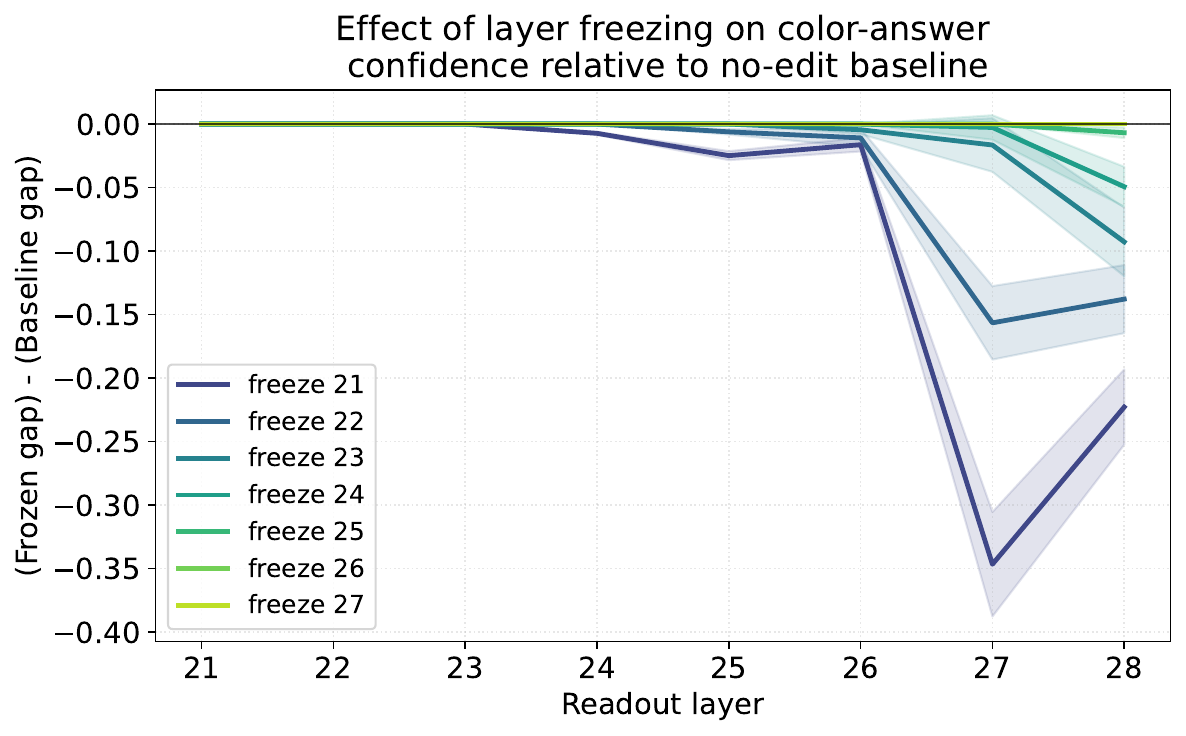}
    \caption{Freezing referent-object vision tokens reduces the VLM's confidence in its final answer. For each layer $i$, referent object hidden states are held fixed beginning at layer $i$. We plot the change in logit gap between the referent and non-referent object's color, compared with the no-freeze baseline. Negative values indicate that the intervention reduces the VLM's confidence in the correct referent attribute. We find that these interventions succeed, implicating later-layer attribute modulation in forming the VLM's final response distributions.}
    \label{fig:freeze_color}
\end{figure}

\section{Conclusion}

We show that goal-directed language induces VLMs to recode their visual representations corresponding to objects.  A linearly decodable reference representation marks goal-relevant objects across counting, yes/no, and spatial tasks in early to mid layers. By extracting the reference representation, we show that this representation causally drives model behavior across multiple tasks, and both synthetic and natural image settings. Beyond denoting relevance, we show that language input selectively amplifies goal-relevant attribute representations across various tasks in later layers, and that this amplification mediates model response distributions. These mechanisms resemble certain aspects of goal-directed recoding of visual representations. Depending on the task, perceptual representations will be modulated in order to support downstream e.g., category discriminations \citep{gazzaley2012top, mckee2014task, freedman2016neuronal}.
Together, our results indicate that linguistic input is used by pretrained VLMs to recode object-level visual representations to support goals or queries that are specified in language.

\section*{Limitations}
Our experiments rely heavily on controlled synthetic datasets with constrained spatial layouts. Our steering vectors are extracted purely from such datasets and our modulation experiments record only how shape and color are modulated instead of other aspects such as size, color, brightness, etc. Additionally, our referent signal may not be the only dominant mechanism. The counting task in particular had lower intervention rates, showing that not all tasks rely entirely on this signal. 

\bibliography{custom}

\clearpage
\appendix

\section{Prompts}
\label{sec:prompts}

\subsection{Synthetic Prompt Template}
\label{sec:synthethic_prompts}

\subsubsection{General Pair Construction}

Each prompt pair is formed by substituting \(\{X\}\) and \(\{Y\}\) into a shared template.

\begin{itemize}
    \item \textbf{Standard pairs:}
    \begin{align*}
        \{X\} &= \texttt{attribute\_present\_in\_image} \\
        \{Y\} &= \texttt{attribute\_not\_present\_in\_image}
    \end{align*}
\item \textbf{NOT pairs:}
\[
\{X\} = \texttt{decision}, \quad
\{Y\} = \texttt{not\_decision}
\]
where \texttt{not\_decision} is the explicit negation of an attribute present in the image (e.g., ``red'' vs.\ ``not red'').
\end{itemize}

\subsubsection{Count}

\begin{enumerate}
\item \texttt{"How many \{X\} shapes are there in the image?"} \\
      \texttt{"How many \{Y\} shapes are there in the image?"}

\item \texttt{"What is the count of \{X\} shapes in the image?"} \\
      \texttt{"What is the count of \{Y\} shapes in the image?"}

\item \texttt{"Identify the number of \{X\} shapes present in the image."} \\
      \texttt{"Identify the number of \{Y\} shapes present in the image."}

\item \texttt{"Determine how many \{X\} shapes appear in this image."} \\
      \texttt{"Determine how many \{Y\} shapes appear in this image."}

\item \texttt{"What is the total number of \{X\} shapes shown?"} \\
      \texttt{"What is the total number of \{Y\} shapes shown?"}
\end{enumerate}

\subsubsection{Yes/No}

\begin{enumerate}
\item \texttt{"Is there a \{X\} shape in the image?"} \\
      \texttt{"Is there a \{Y\} shape in the image?"}

\item \texttt{"Does the image contain any \{X\} shape?"} \\
      \texttt{"Does the image contain any \{Y\} shape?"}

\item \texttt{"Can you find a \{X\} shape in the picture?"} \\
      \texttt{"Can you find a \{Y\} shape in the picture?"}

\item \texttt{"Is a \{X\} shape present in this image?"} \\
      \texttt{"Is a \{Y\} shape present in this image?"}

\item \texttt{"Do you see any \{X\} shape in the image?"} \\
      \texttt{"Do you see any \{Y\} shape in the image?"}
\end{enumerate}

\subsubsection{Spatial}

Let \(r \in \{\texttt{left of}, \texttt{right of}, \texttt{above}, \texttt{below}\}\), and let \(\bar{r}\) denote its opposite.

\begin{align*}
\texttt{decision\_text} &= \parbox[t]{4.5cm}{\raggedright \texttt{"\{rel(r)\} the \{target\_color\} \{target\_shape\}"}} \\[1.5ex]
\texttt{opposite\_text} &= \parbox[t]{4.5cm}{\raggedright \texttt{"\{rel(}$\bar{\texttt{r}}$\texttt{)\} the \{target\_color\} \{target\_shape\}"}}
\end{align*}

These are inserted into the following templates:

\begin{enumerate}
\item \texttt{"What shape is \{T\}?"}
\item \texttt{"Which shape is \{T\}?"}
\item \texttt{"Identify the shape that is \{T\}."}
\item \texttt{"List the shape that is \{T\}."}
\item \texttt{"Find the shape that is \{T\}."}
\item \texttt{"Select the shape that is \{T\}."}
\item \texttt{"The shape located \{T\} is which?"}
\item \texttt{"What object is \{T\}?"}
\item \texttt{"Which object lies \{T\}?"}
\item \texttt{"Describe the shape positioned \{T\}."}
\end{enumerate}

Each template \(t(\cdot)\) produces a contrastive pair:
\[
\big(t(\texttt{decision\_text}), \; t(\texttt{opposite\_text})\big)
\]

\subsection{Probing Prompts}
\label{sec:probe_prompts}
For the reference prompts we only use the first prompt in each of the pairs described above. Empty prompt is an empty string ``''. Distractor caption prompts are of the form of ``An image of {color 1}{shape 1} and {color 2}{shape 2} and {color 3}{shape 3}.'' where none of the colors and shapes exist in the image. All caption prompts are of the same form as distractor caption prompts except each {color i}{shape i} describes an existing object.

\subsection{Synthetic Intervention Prompts}
\label{sec:synthethic_intervention_prompts}
\begin{itemize}
\item \textbf{Count:} \\
\texttt{"How many \{decision\} shapes are there in the image? Answer with a number."}

\item \textbf{Yes/No:} \\
\texttt{"Is there a \{attr\} shape in the image? Answer with yes or no."}

\item \textbf{Spatial:} \\
\texttt{"Name only the shape and color of the object \{relation\} of the \{target\_color\} \{target\_shape\}. Be concise."}
\end{itemize}

\subsection{Natural Intervention Prompts}
\label{sec:natural_intervention_prompts}
\begin{itemize}
\item \textbf{Count:} \\
\texttt{"How many \{target\_object\} are in the image? Answer with a single number."}

\item \textbf{Yes/No (present):} \\
\texttt{"Is there a \{target\_object\} in the image? Answer with yes or no."}

\item \textbf{Yes/No (absent):} \\
\texttt{"Is there a \{absent\_object\} in the image? Answer with yes or no."}

\item \textbf{Spatial-left:} \\
\texttt{"Name all objects to the left of the \{middle\_object\}."}

\item \textbf{Spatial-right:} \\
\texttt{"Name all objects to the right of the \{middle\_object\}."}
\end{itemize}

For the spatial task a correct answer is one where both the baseline answer is not mentioned and the new object is mentioned.

\subsection{Modulation Prompts}
\label{sec:priming_prompts}
\subsubsection{Front vs.\ Back Modulation}

\begin{itemize}
\item \texttt{"Describe only the shape that is in front of another shape."}
\item \texttt{"Describe only the shape that is behind another shape."}
\end{itemize}

\subsubsection{Shape vs.\ Color Modulation}

\begin{itemize}
\item \texttt{"Focus on the shape of each object in the image."}
\item \texttt{"Focus on the color of each object in the image."}
\end{itemize}

\subsubsection{Referent Object Modulation}

\begin{itemize}
\item \texttt{"What objects are \{relation\} of the \{target\_color\} \{target\_shape\}?"}
\item \texttt{"What objects are \{opposite\_relation\} of the \{target\_color\} \{target\_shape\}?"}
\end{itemize}

\section{Qwen Extra Probing Experiments}
\label{sec:qwen_probe_extra}
\begin{figure}[t]
  \centering
  \includegraphics[width=\columnwidth]{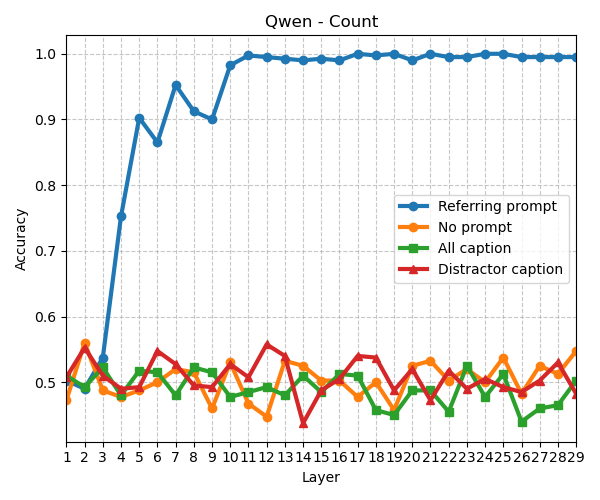}
  
  \includegraphics[width=\columnwidth]{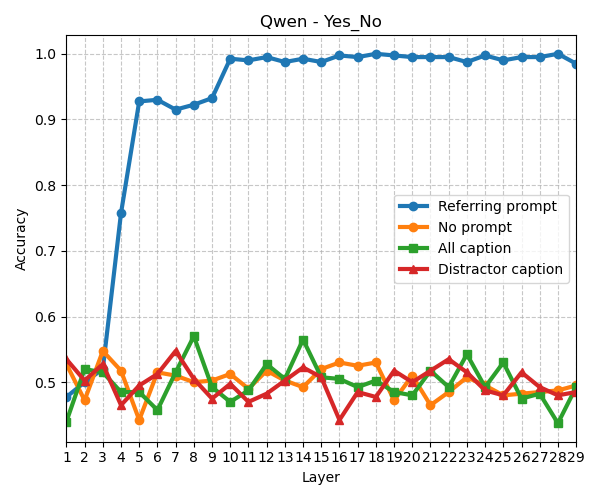}
  
  \caption{Probing analysis of Qwen for Counting and Yes/No task.}
  \label{fig:qwen_probing}
\end{figure}

We report the extra Counting and Yes/No probing synthetic experiments for Qwen in Figure~\ref{fig:qwen_probing}.

\section{Qwen Natural Cross Task Interventions}
\label{sec:cross task_intervention}
\begin{figure}[t]
  \centering
  \includegraphics[width=\columnwidth]{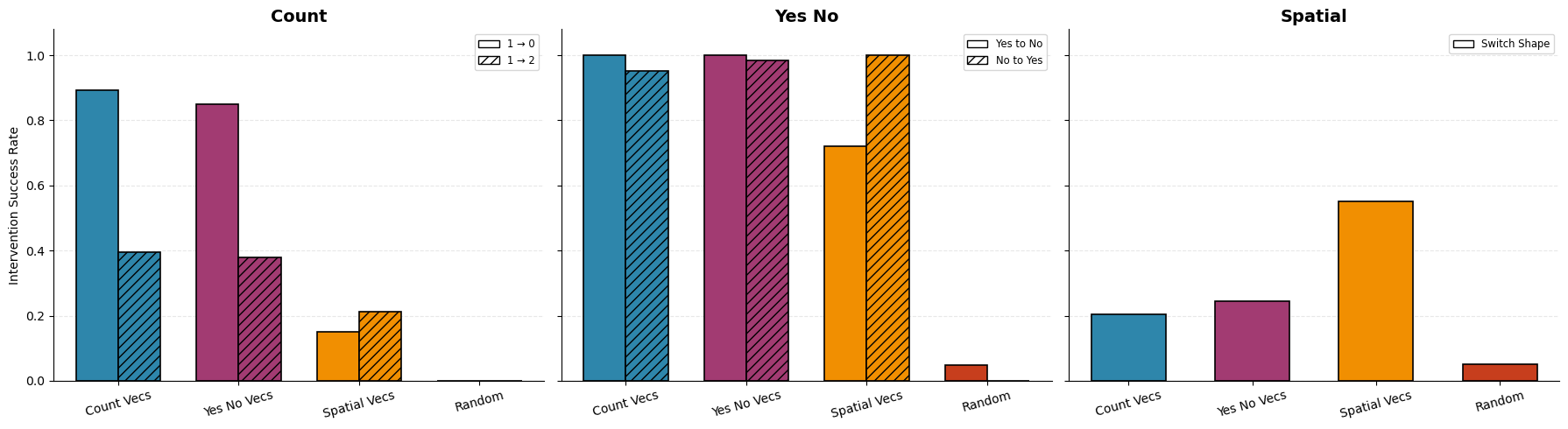}
  
  \caption{Qwen cross task intervention success rates on the natural dataset. Reference vectors consistently achieve high success rate over random vector controls.}
  \label{fig:qwen_natural_cross_task}
\end{figure}

We report the cross task intervention success rates on the natural dataset in Figure~\ref{fig:qwen_natural_cross_task}. Same task vectors achieve the highest success rates, and while cross task vectors don't achieve as high, they all greatly outperform the random vector controls. These results are consistent with the intervention experiments on the synthetic dataset.

\section{InternVL3 experiments}
\label{sec:internvl3_experiments}

\begin{figure}[!t]
  \centering
  \includegraphics[width=\columnwidth]{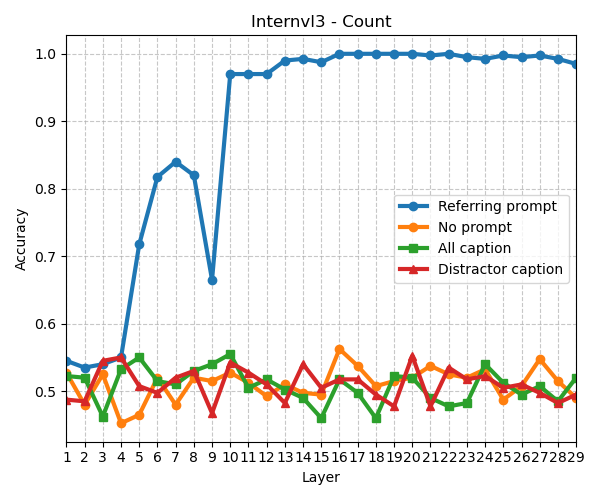}
  
  \includegraphics[width=\columnwidth]{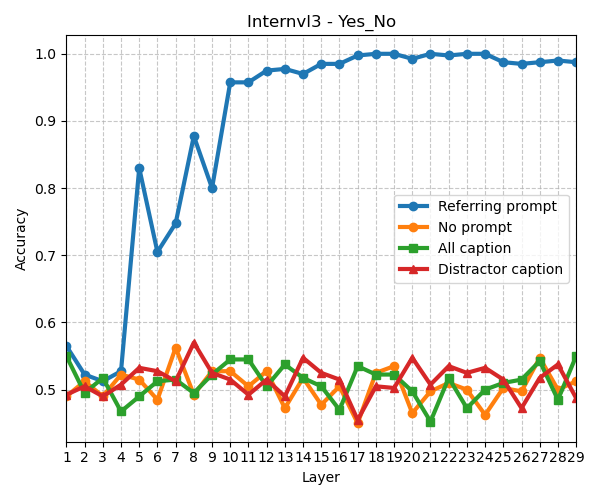}
  
  \includegraphics[width=\columnwidth]{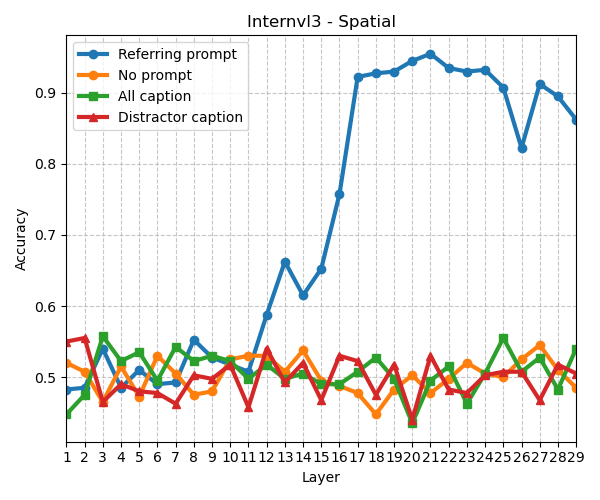}
  
  \caption{Probing analysis of InternVL for Counting, Yes/No, and Spatial task.}
  \label{fig:intern_probing}
\end{figure}

\begin{figure}[!t]
  \centering
  \includegraphics[width=\columnwidth]{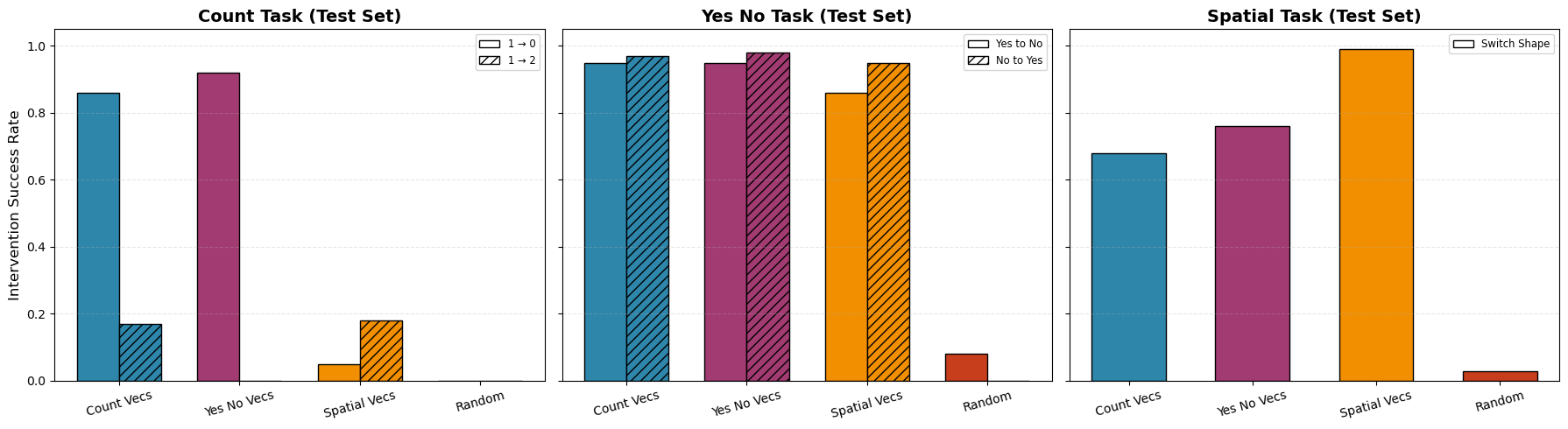}
  \caption{Synthetic interventions}
  \label{fig:intern_synthethic}
\end{figure}

\begin{figure}[!t]
  \centering
  \includegraphics[width=\columnwidth]{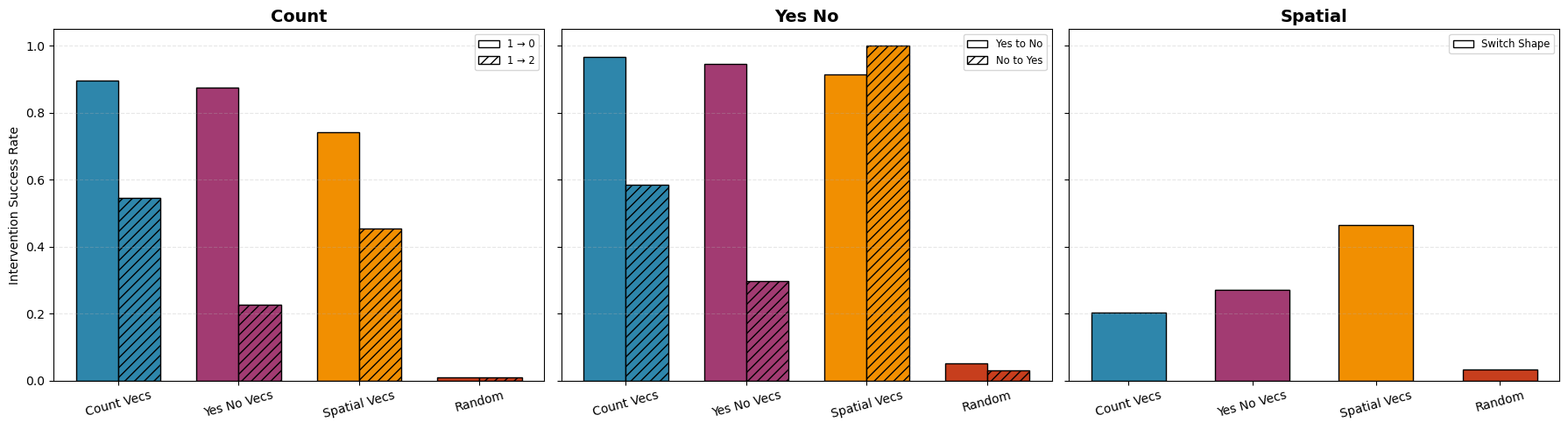}
  \caption{Natural interventions}
  \label{fig:intern_natural}
\end{figure}

\begin{figure}[!t]
  \centering
  \includegraphics[width=\columnwidth]{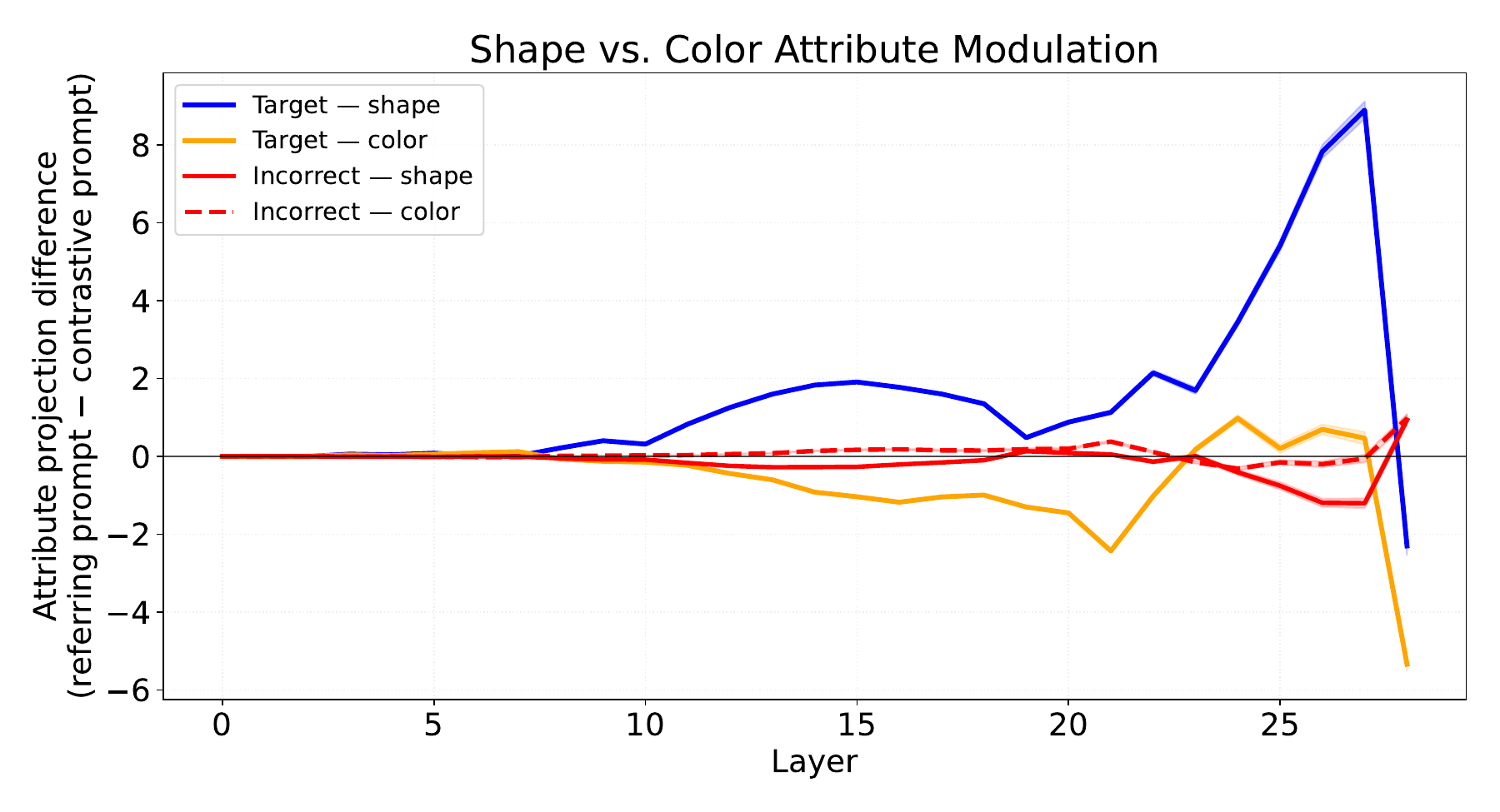}
  \caption{Shape vs.\ color modulation}
  \label{fig:intern_color_shape_concept_priming}
\end{figure}

\begin{figure}[!t]
  \centering
  \includegraphics[width=\columnwidth]{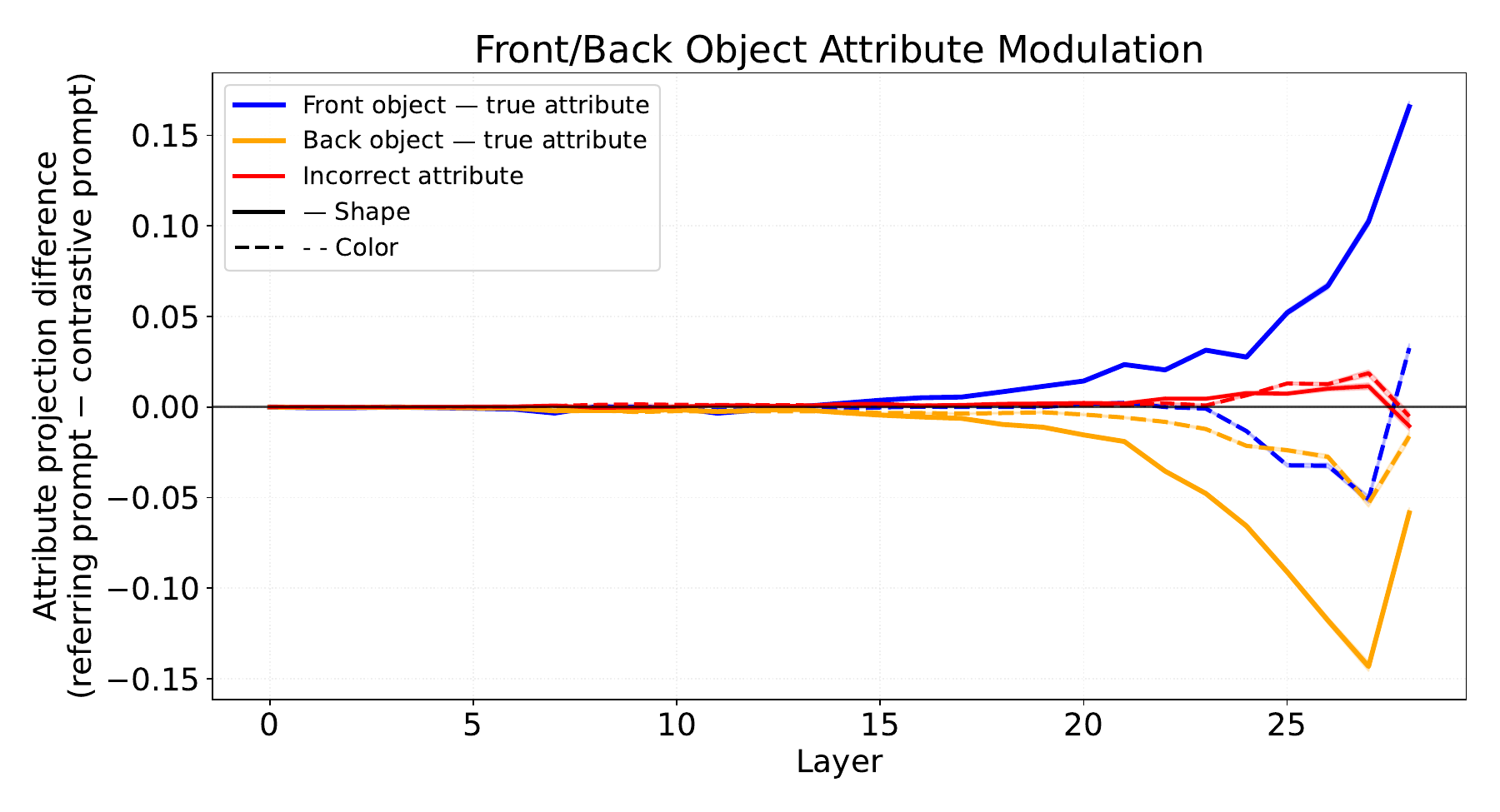}
  \caption{Front vs.\ back modulation}
  \label{fig:intern_front_back_concept_priming}
\end{figure}

\begin{figure}[!t]
  \centering
  \includegraphics[width=\columnwidth]{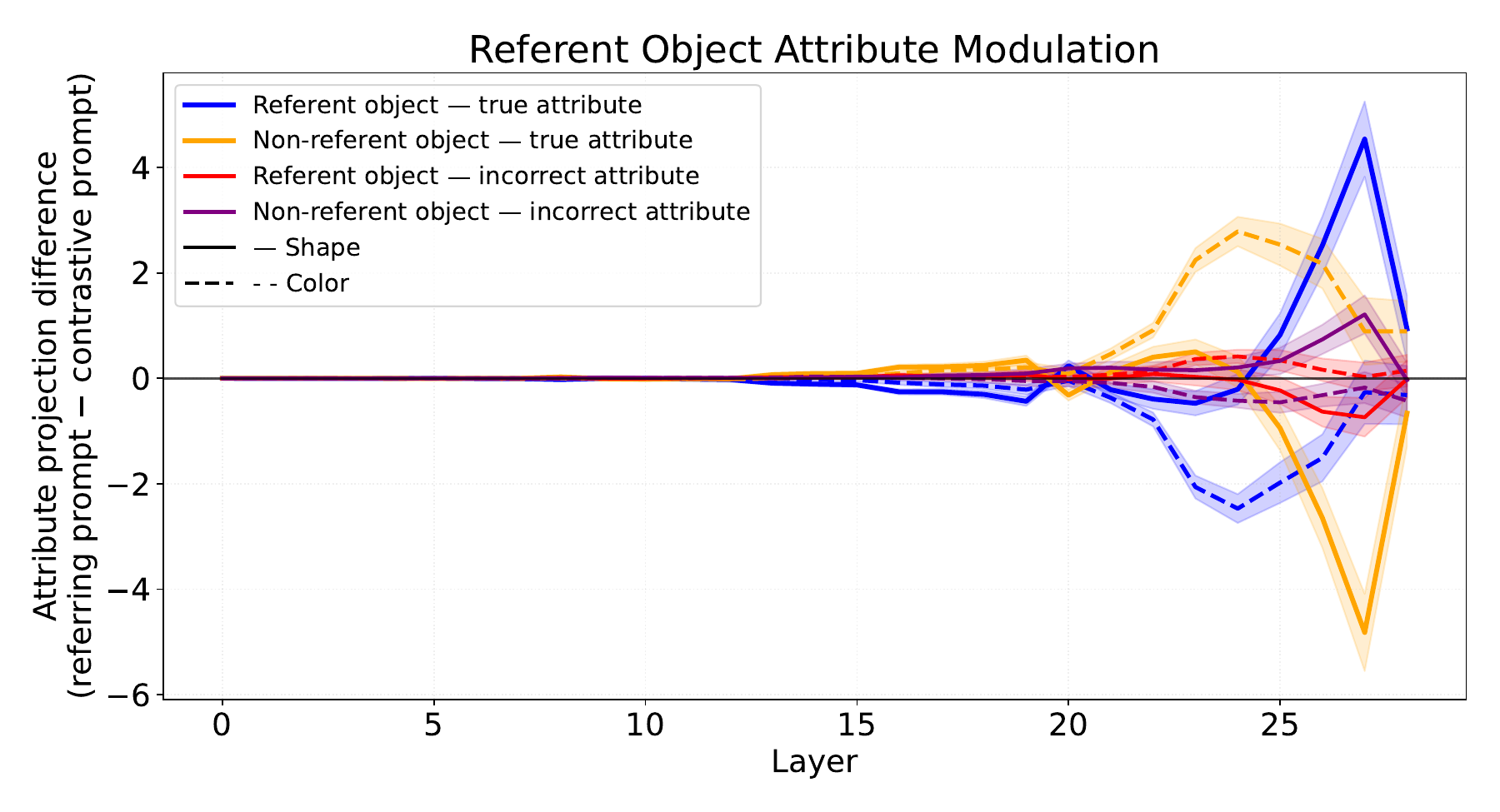}
  \caption{Referent object modulation}
  \label{fig:intern_referred_concept_priming}
\end{figure}

We replicate the probing \ref{fig:intern_probing}, synthetic intervention \ref{fig:intern_synthethic}, natural intervention \ref{fig:intern_natural}, shape and color modulation \ref{fig:intern_color_shape_concept_priming}, front and back modulation \ref{fig:intern_front_back_concept_priming}, and referent object modulation \ref{fig:intern_referred_concept_priming}. Unlike Qwen, we find that the steering vector calculation for the counting and yes/no tasks, using both the standard and NOT pair constructions, results in subpar performance. Therefore, we only use standard pairs for these tasks. The baseline accuracy during the interventions for the counting, Yes/No, and spatial tasks is around 100, 92, and 98\%, respectively. Additionally, for the natural intervention tasks, Qwen allows dynamically sized images as input, meaning it does not automatically resize images to a fixed resolution. InternVL3 however, does. Therefore, for these experiments, all images are resized to 448 $\times$ 448 px.

\section{Steering Hyperparameters}
\label{sec:hyperparameters}
\begin{figure*}[!t]
  \centering

  \textbf{Qwen2.5-VL-7B-Instruct — Synthetic}

  \vspace{0.3em}

  \includegraphics[width=0.32\textwidth]
  {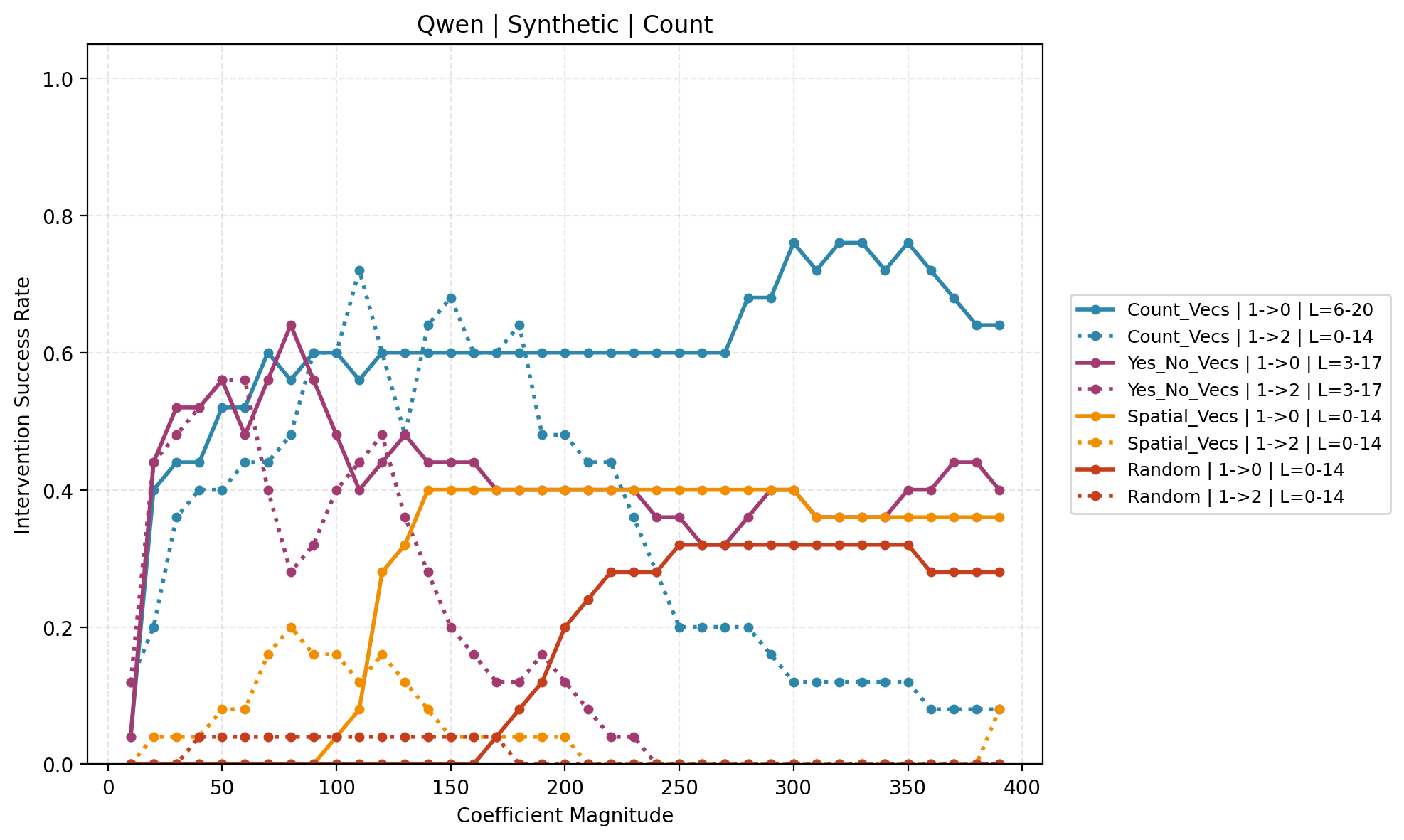}
  \hfill
  \includegraphics[width=0.32\textwidth]
  {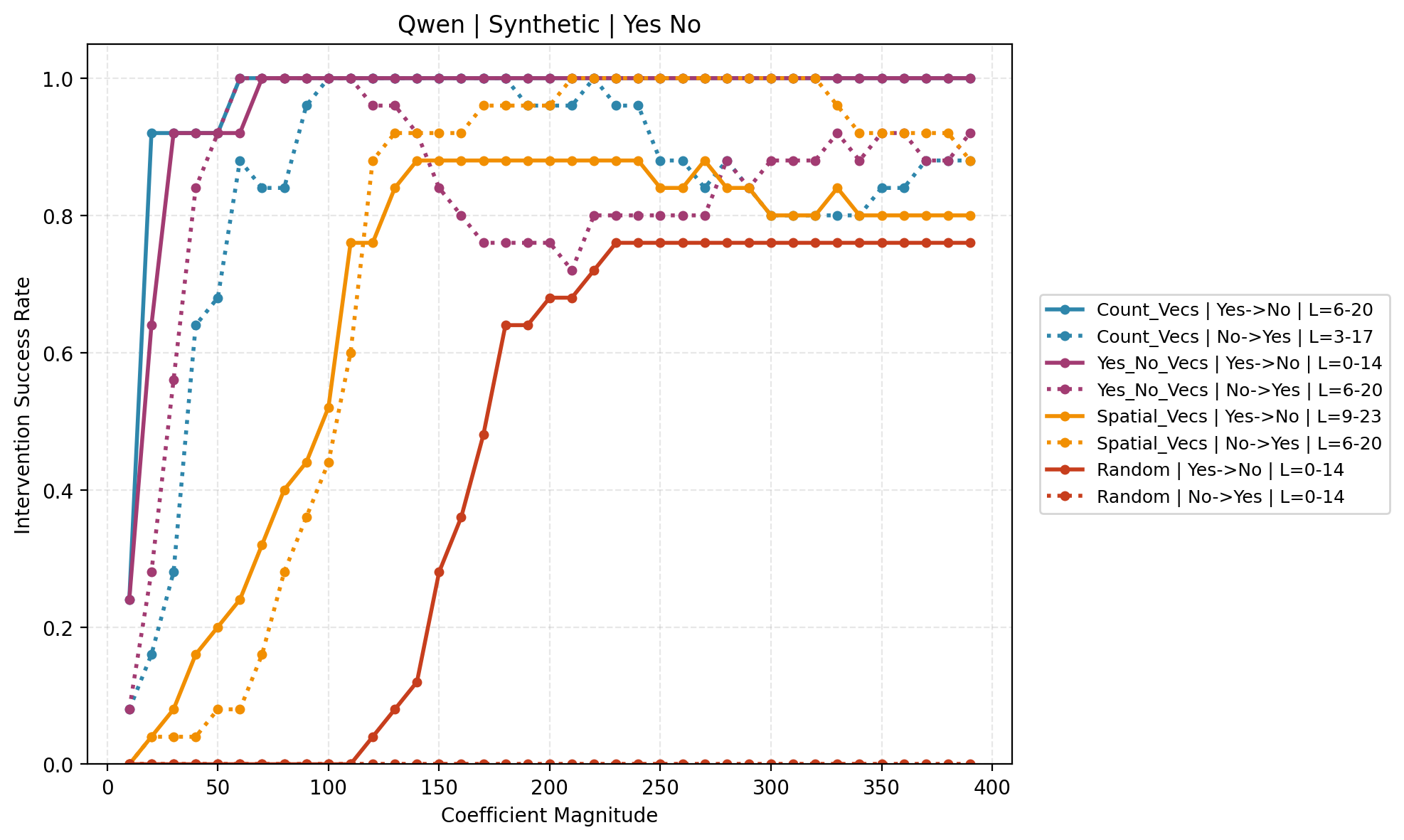}
  \hfill
  \includegraphics[width=0.32\textwidth]
  {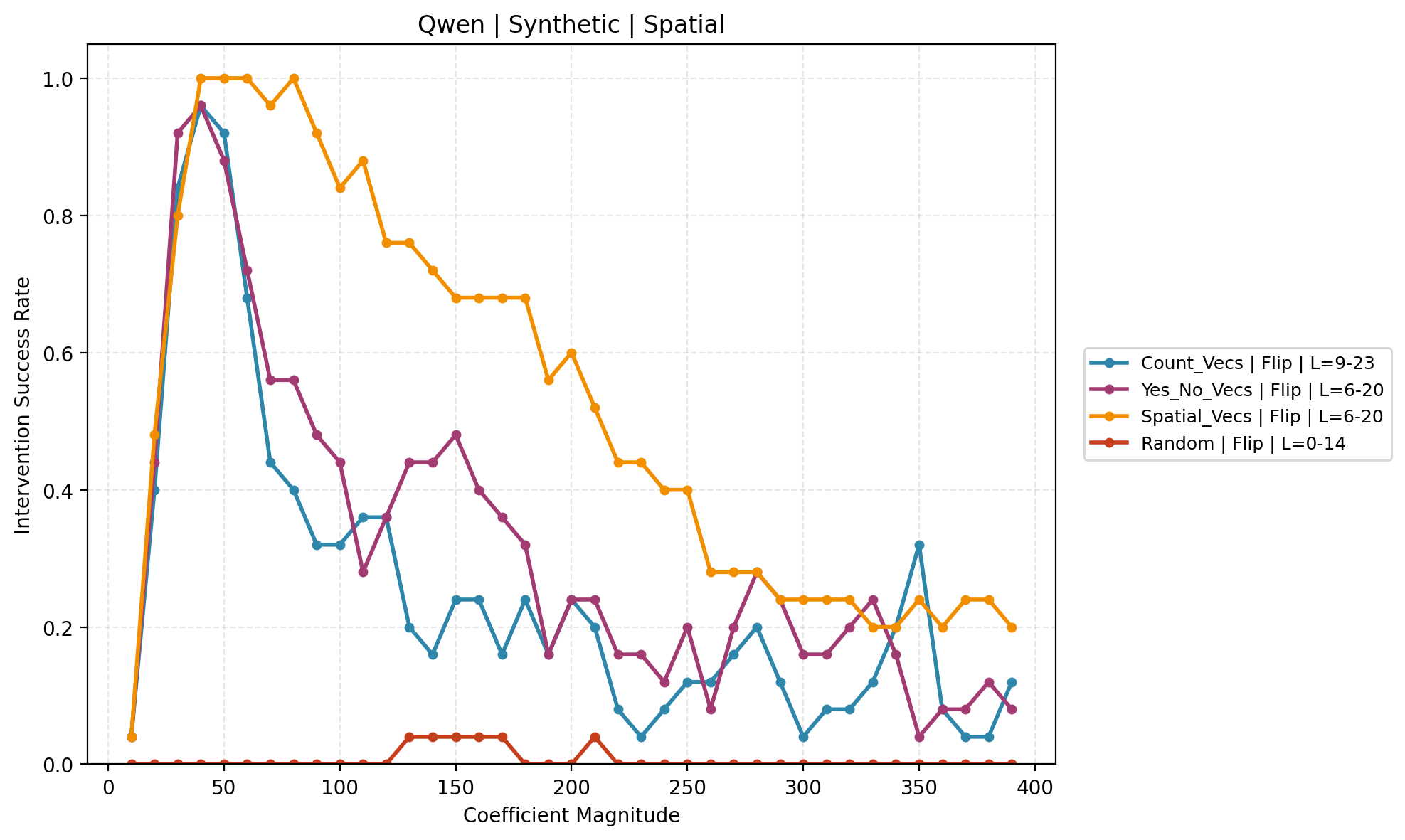}

  \vspace{0.7em}

  \textbf{Qwen2.5-VL-7B-Instruct — Natural}

  \vspace{0.3em}

  \includegraphics[width=0.32\textwidth]
  {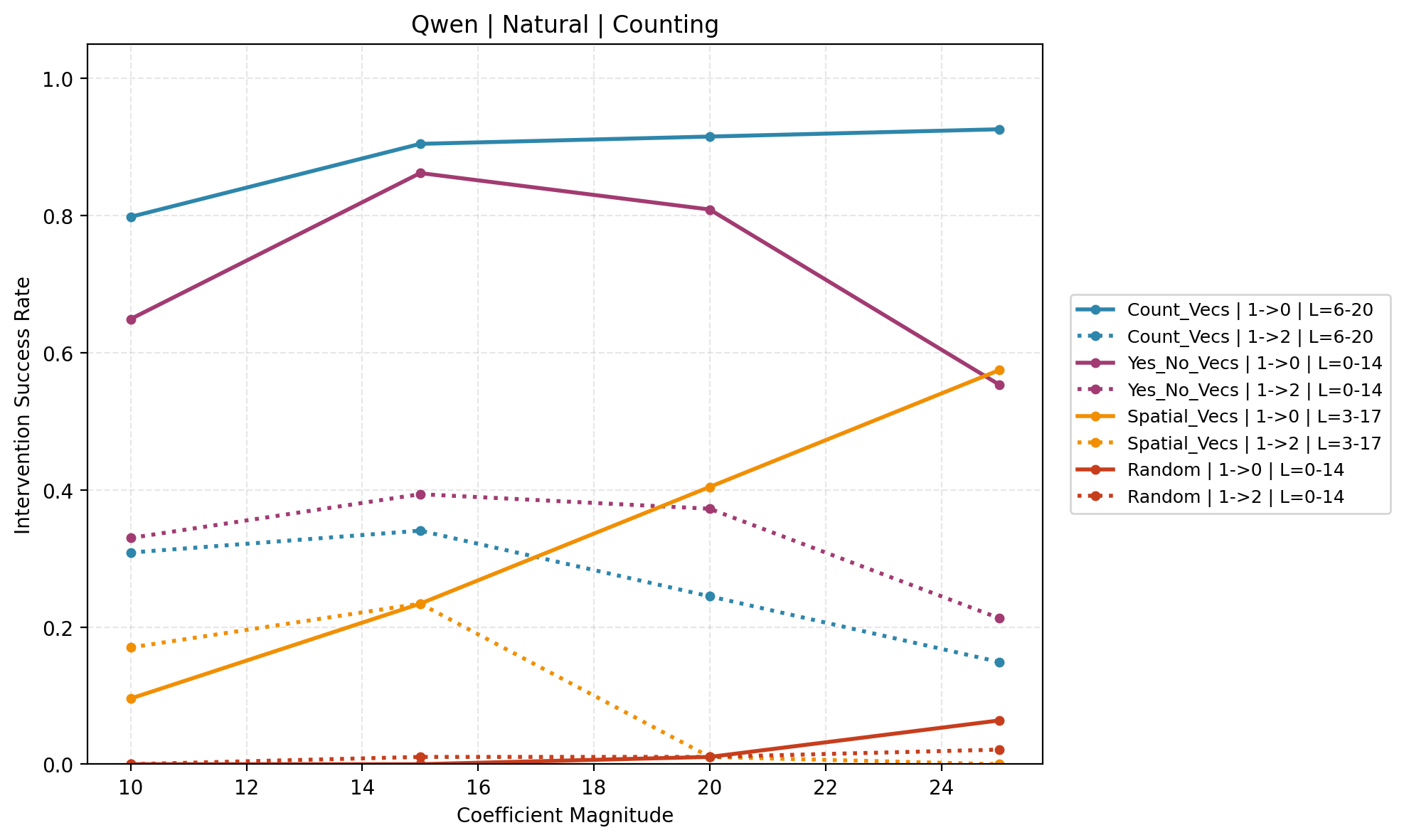}
  \hfill
  \includegraphics[width=0.32\textwidth]
  {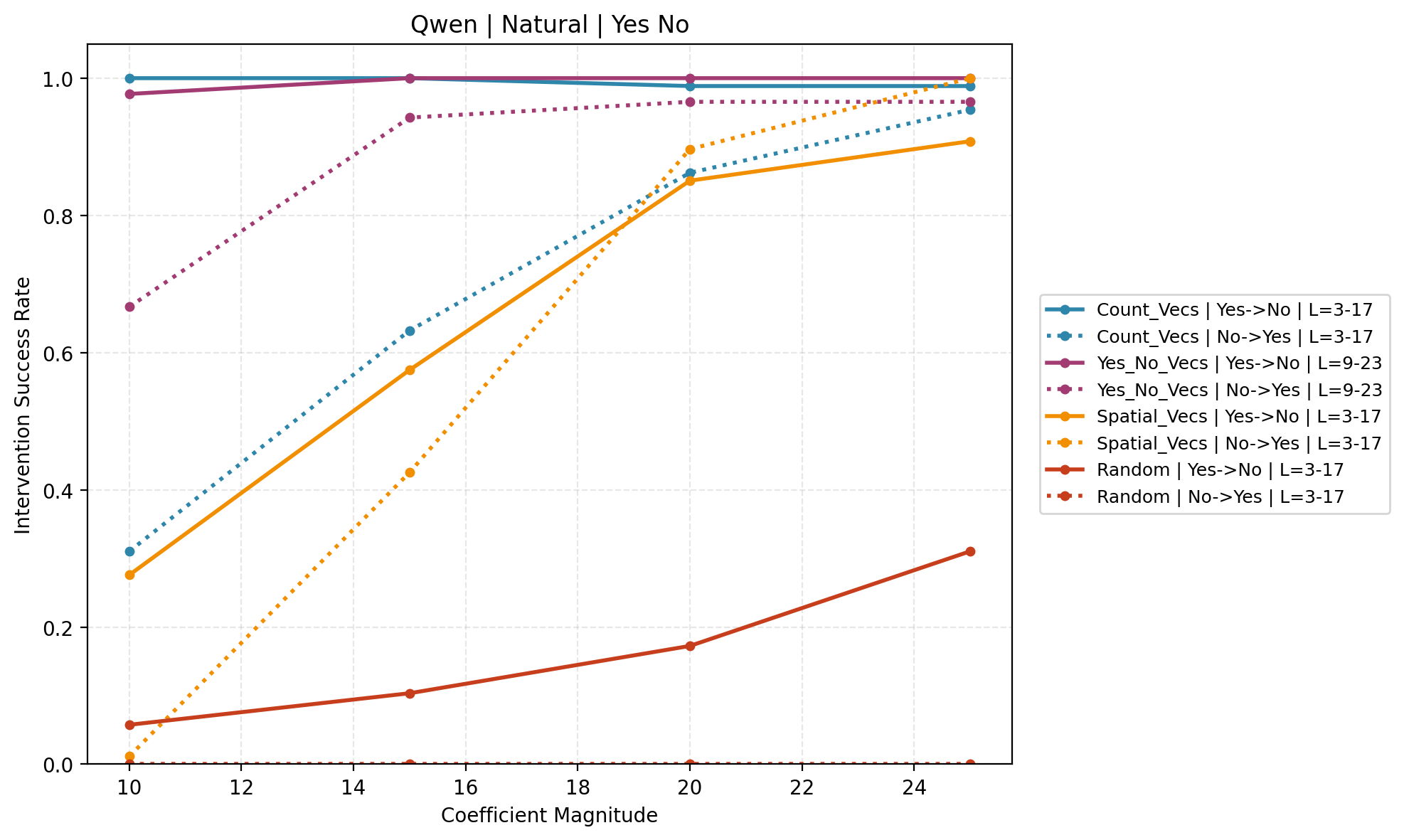}
  \hfill
  \includegraphics[width=0.32\textwidth]
  {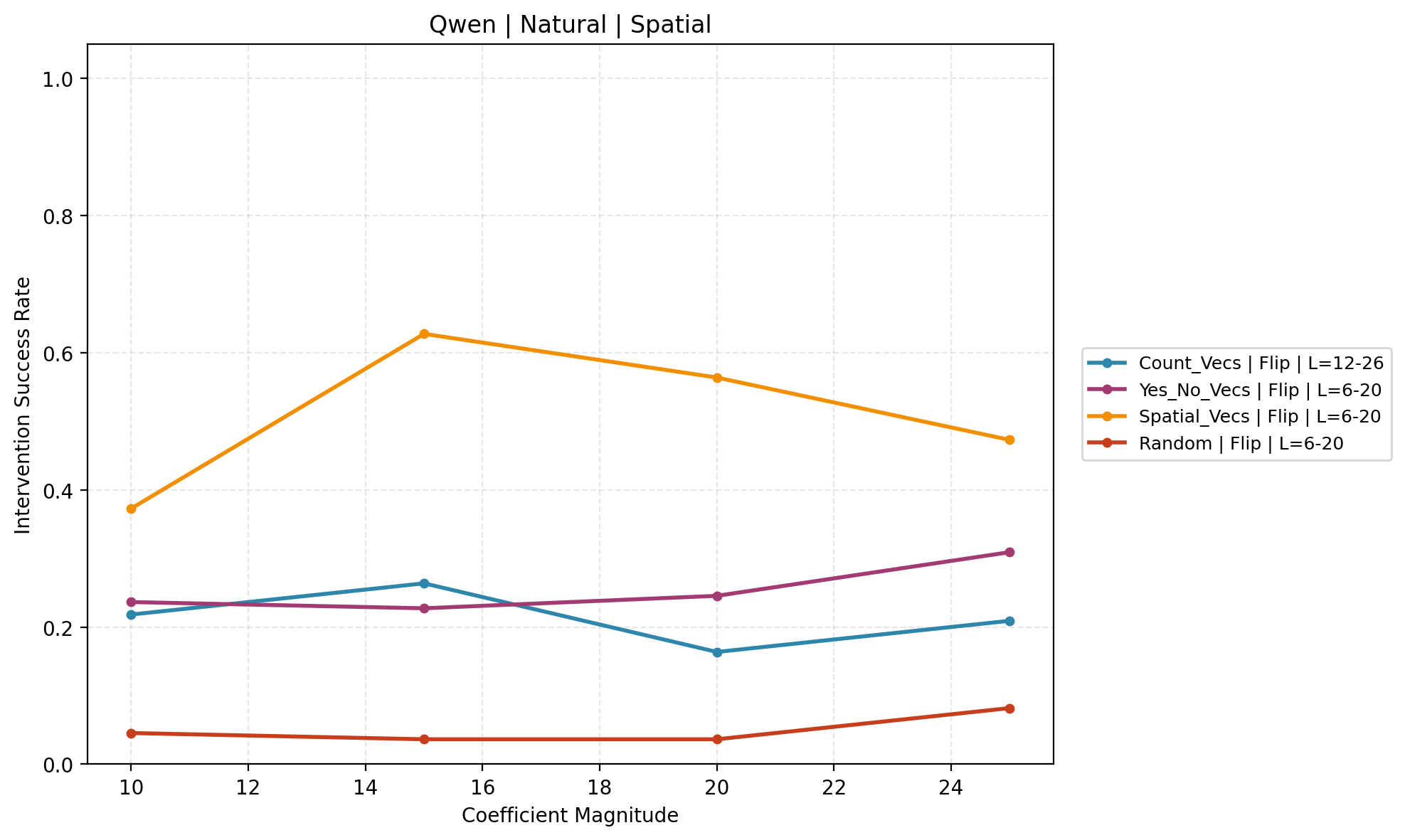}

  \vspace{1em}

  \textbf{InternVL3-8B — Synthetic}

  \vspace{0.3em}

  \includegraphics[width=0.32\textwidth]
  {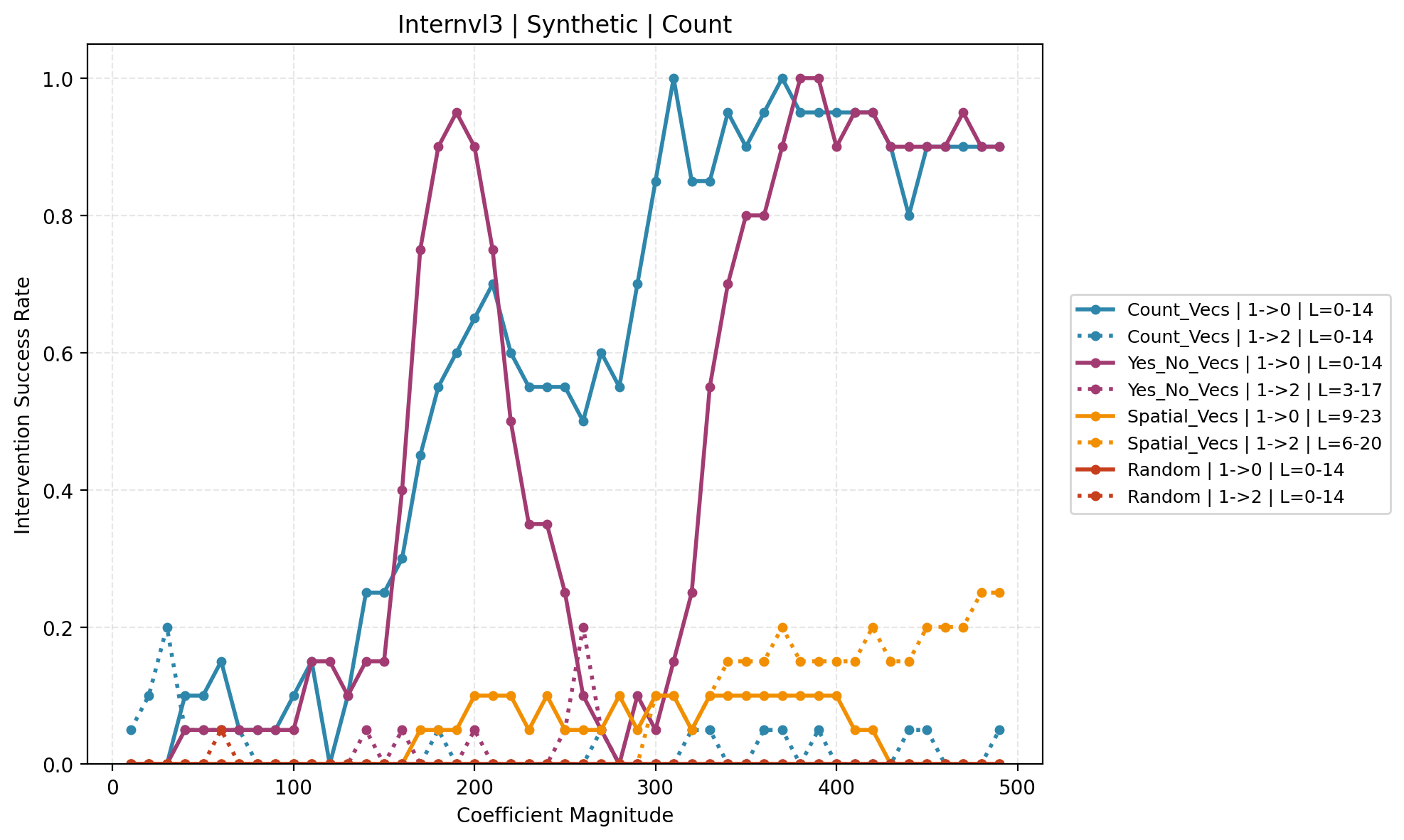}
  \hfill
  \includegraphics[width=0.32\textwidth]
  {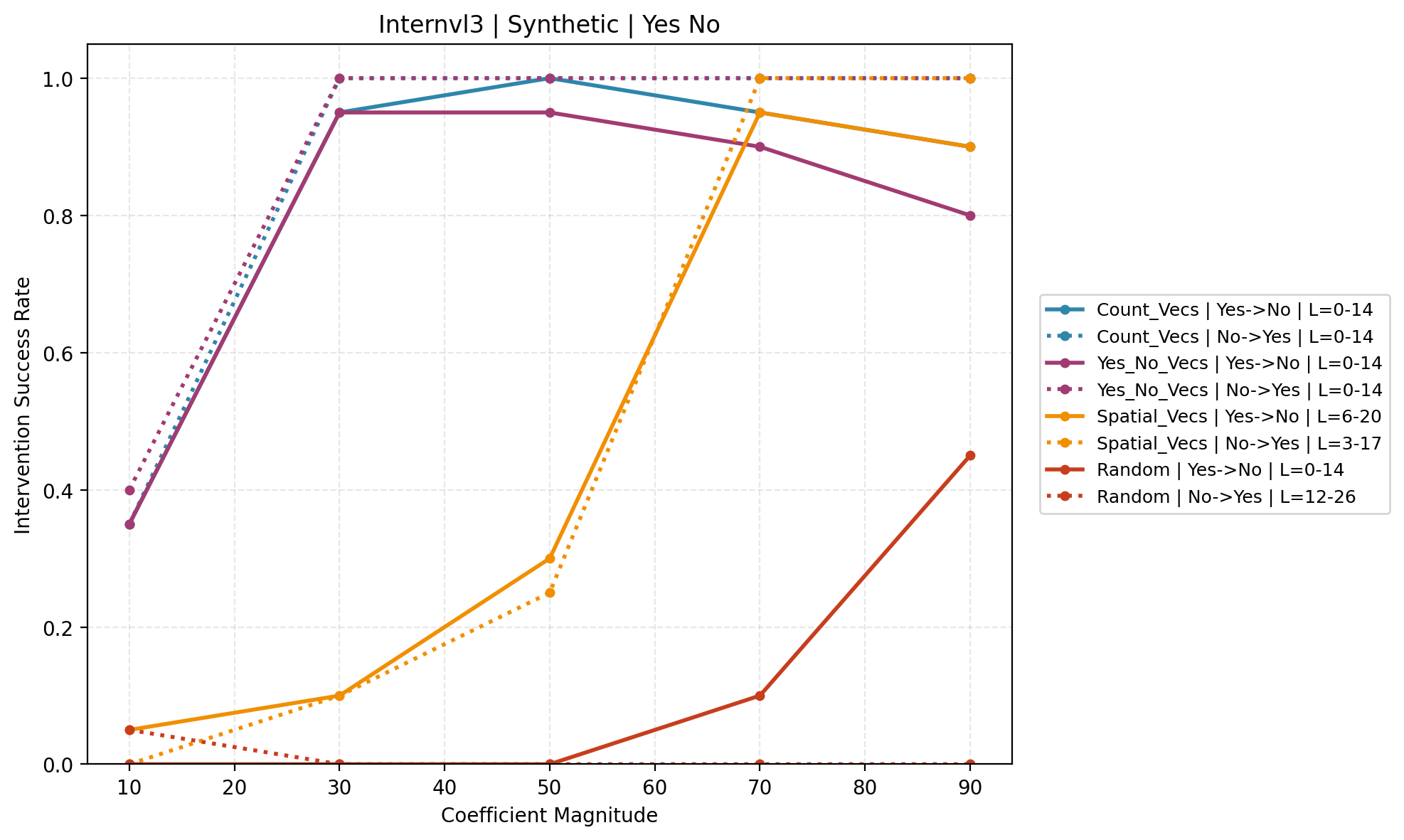}
  \hfill
  \includegraphics[width=0.32\textwidth]
  {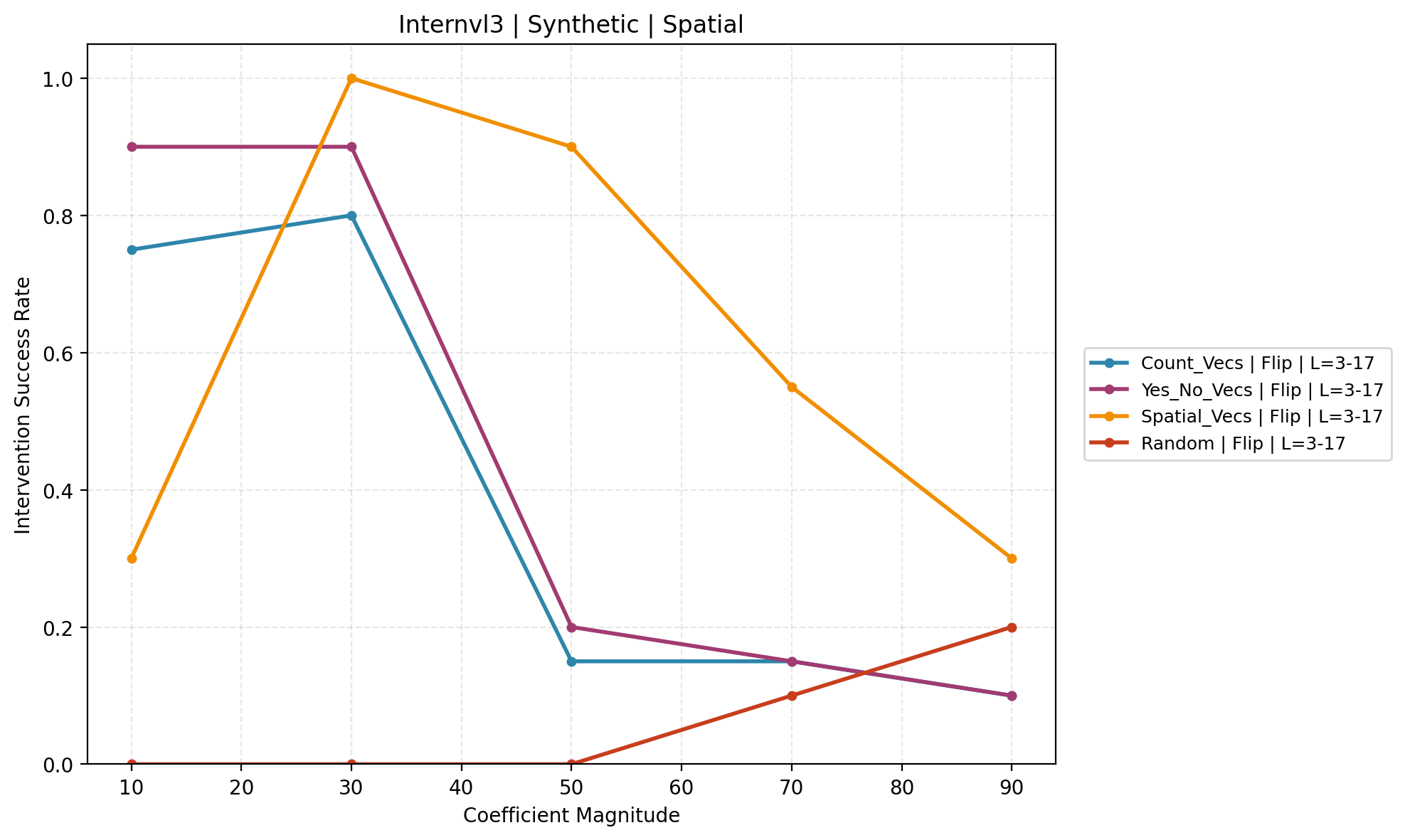}

  \vspace{0.7em}

  \textbf{InternVL3-8B — Natural}

  \vspace{0.3em}

  \includegraphics[width=0.32\textwidth]
  {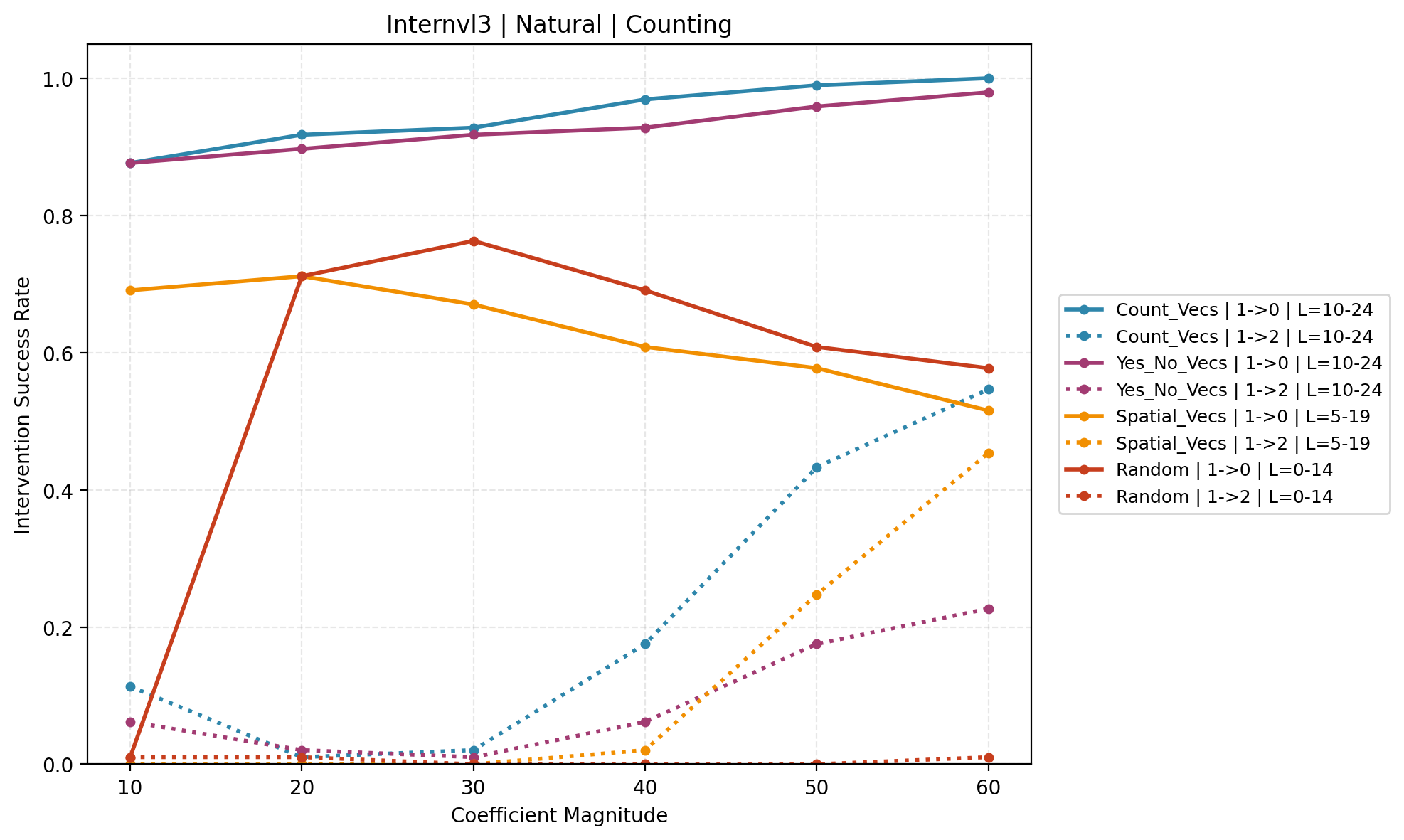}
  \hfill
  \includegraphics[width=0.32\textwidth]
  {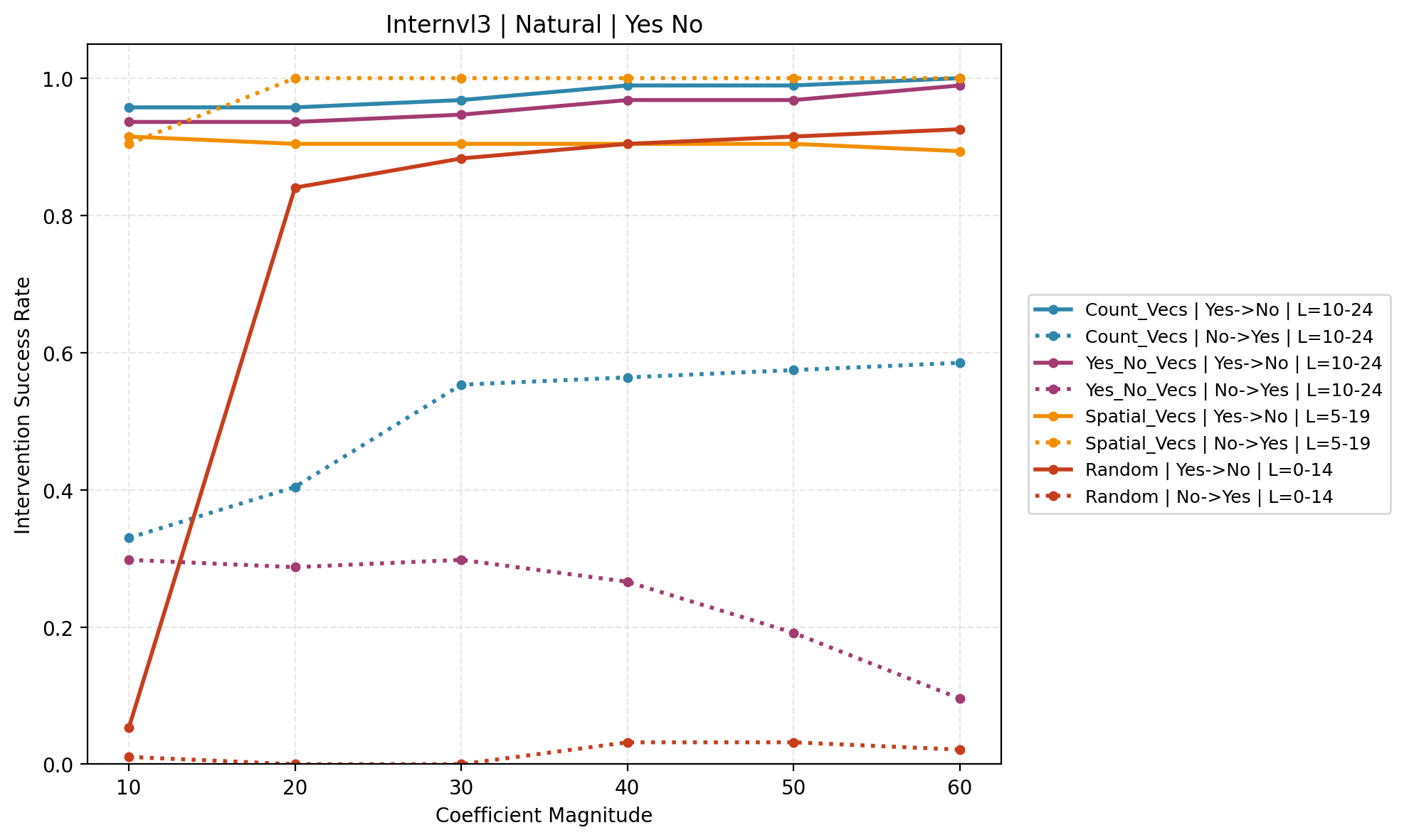}
  \hfill
  \includegraphics[width=0.32\textwidth]
  {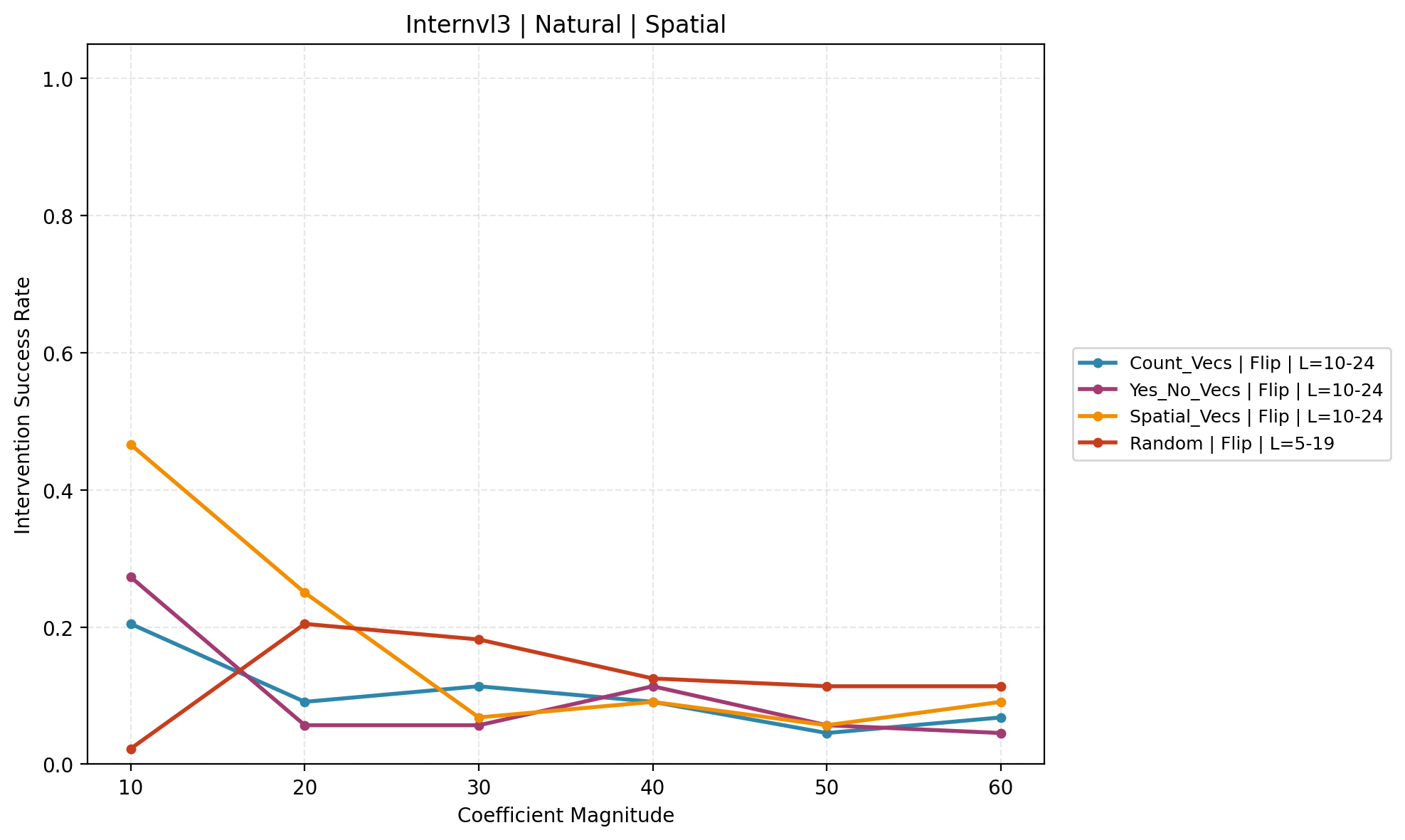}

  \caption{
  Steering coefficient sweeps across models, datasets, and tasks.
  Rows correspond to synthetic and natural evaluations for Qwen2.5-VL-7B-Instruct and InternVL3-8B.
  }
  \label{fig:hyperparams}
\end{figure*}

We report intervention success rates as a function of coefficient magnitude for synthetic and natural datasets in Figure~\ref{fig:hyperparams}. We record the coefficient and layer range that gave the highest intervention rate for each task specific or random steering vector. We then fix the layer range and record the intervention rate at different coefficients. All results are from the validation set. Notably, random vectors can have modest performance on interventions that can benefit from destroying representations such as Counting 1 $\rightarrow$ 0 and Yes/No Yes $\rightarrow$ No but require larger coefficients than task specific reference vectors to achieve similar performance. Therefore, for evaluation, we mark the coefficient at which random vectors start to succeed, then prevent any other steering vectors from using that coefficient or higher. 

\section{Formal Definitions for Attribute Modulation}
\label{sec:appendix_modulation}

\subsection{Concept Vector Construction}
\label{sec:appendix_concept_vectors}

Recall from Section~\ref{sec:methods} that in the synthetic dataset each object
occupies a $4\times4$ patch of vision tokens, so we write object hidden states as
$h_l(o,p)\in\mathbb{R}^{4\times4\times d}$, where $o$ denotes the object,
$p$ the prompt, and $l$ the layer.

For concept $a$ at layer $l$, we define the concept vector
\begin{equation}
    V_l^a =
    \frac{
    \mu_l^a - \mu_l^{\mathrm{glob}}
    }{
    \|\mu_l^a - \mu_l^{\mathrm{glob}}\|
    },
\end{equation}
where $\mu_l^a\in\mathbb{R}^{4\times4\times d}$ is the mean hidden-state tensor
over all object patches containing concept $a$, and
$\mu_l^{\mathrm{glob}}$ is the mean over all object patches. Consequently, $V_l^a\in\mathbb{R}^{4\times4\times d}$ contains a $d$-dimensional direction for each of the 16 relative spatial positions. To measure the degree to which an object representation encodes a concept, we take the dot product between each vision token hidden state and its corresponding vector in $V_l^a$ then average those values:
\begin{equation}
    \mathrm{Proj}(o,a,p,l)
    =
    \frac{1}{16}
    \sum_{u=1}^{4}
    \sum_{v=1}^{4}
    h_l(o,p)_{u,v}^{\top}
    V_l^a{}_{u,v}.
    \label{eq:proj_appendix}
\end{equation}

A higher projection indicates that a concept, e.g. "Red" or "Square", is more present in the object. Throughout the modulation experiments we also
compute projections onto absent concepts --- randomly chosen shapes or colors
that do not appear anywhere in the image --- as controls.

\subsection{Explicit Shape and Color Modulation}
\label{sec:appendix_shape_color_modulation}
To visualize attribute-specific shifts relative to the baseline description
prompt, we compute
\begin{align}
\Delta_{\mathrm{shape}}^{\mathrm{base}}(o,l)
&=
\mathrm{Proj}(o,s(o),p_{\mathrm{shape}},l)
\nonumber \\
&\quad -
\mathrm{Proj}(o,s(o),p_{\mathrm{base}},l), \\
\Delta_{\mathrm{color}}^{\mathrm{base}}(o,l)
&=
\mathrm{Proj}(o,c(o),p_{\mathrm{color}},l)
\nonumber \\
&\quad -
\mathrm{Proj}(o,c(o),p_{\mathrm{base}},l),
\end{align}
where $s(o)$ denotes the shape concept vector of object $o$, $c(o)$ denotes the color concept vector of object $o$, and $p_{\mathrm{base}}$ denotes the generic description prompt that lists
the attributes of all objects in the image without explicitly emphasizing
shape or color.

\subsection{Front/Back Object Modulation}
\label{sec:appendix_front_back_modulation}

For the overlapping-object dataset, each object occupies the full image which is a 
$16\times16$ region of vision tokens. To remain consistent with the
projection computation in Equation~\ref{eq:proj_appendix}, we divide each
object region into sixteen non-overlapping $4\times4$ windows and average
their projections.

Let $\mathcal{W}$ denote the set of all $4\times4$ windows belonging to
object $o$. We define the averaged projection
\begin{equation}
    \widetilde{\mathrm{Proj}}(o,a,p,l)
    =
    \frac{1}{|\mathcal{W}|}
    \sum_{w\in\mathcal{W}}
    \mathrm{Proj}(o_w,a,p,l),
\end{equation}
where $o_w$ denotes the object sub-patch corresponding to window $w$.

Given a front-focused prompt $p^+$ and a back-focused prompt $p^-$,
the modulation score is
\begin{equation}
    \Delta_{\mathrm{front}}(o,a,l)
    =
    \widetilde{\mathrm{Proj}}(o,a,p^+,l)
    -
    \widetilde{\mathrm{Proj}}(o,a,p^-,l).
\end{equation}

We compute these modulation values using the concept vector $a$ corresponding to the foreground object's true shape/color, the background object's true shape/color, and absent
shape/color controls.

\subsection{Referent Object Modulation}
\label{sec:appendix_referent_modulation}

For the referent modulation experiments, we compare a prompt pair
$(p^+, p^-)$ such that $p^+$ refers to object $o^+$ while $p^-$ refers to
object $o^-$. For example, $p^+$ may ask ``What object is left of the blue
square?'' while $p^-$ asks ``What object is right of the blue square?''

We measure modulation at both object positions. Referent-position modulation
is defined as
\begin{align}
\Delta_{\mathrm{ref}}(a,l)
&=
\mathrm{Proj}(o^+,a,p^+,l)
\nonumber \\
&\quad -
\mathrm{Proj}(o^+,a,p^-,l),
\end{align}
while non-referent-position modulation is
\begin{align}
\Delta_{\mathrm{nonref}}(a,l)
&=
\mathrm{Proj}(o^-,a,p^+,l)
\nonumber \\
&\quad -
\mathrm{Proj}(o^-,a,p^-,l).
\end{align}

Here, $a$ may correspond either to the concept vector representing the true shape/color of the specified object or to an absent-concept control.

\section{Color vs.\ shape modulation extra experiments}
\subsection{Scatterplot analysis}
\label{sec:scatterplot}
\begin{figure*}[!t]
    \centering
    \includegraphics[width=\textwidth]{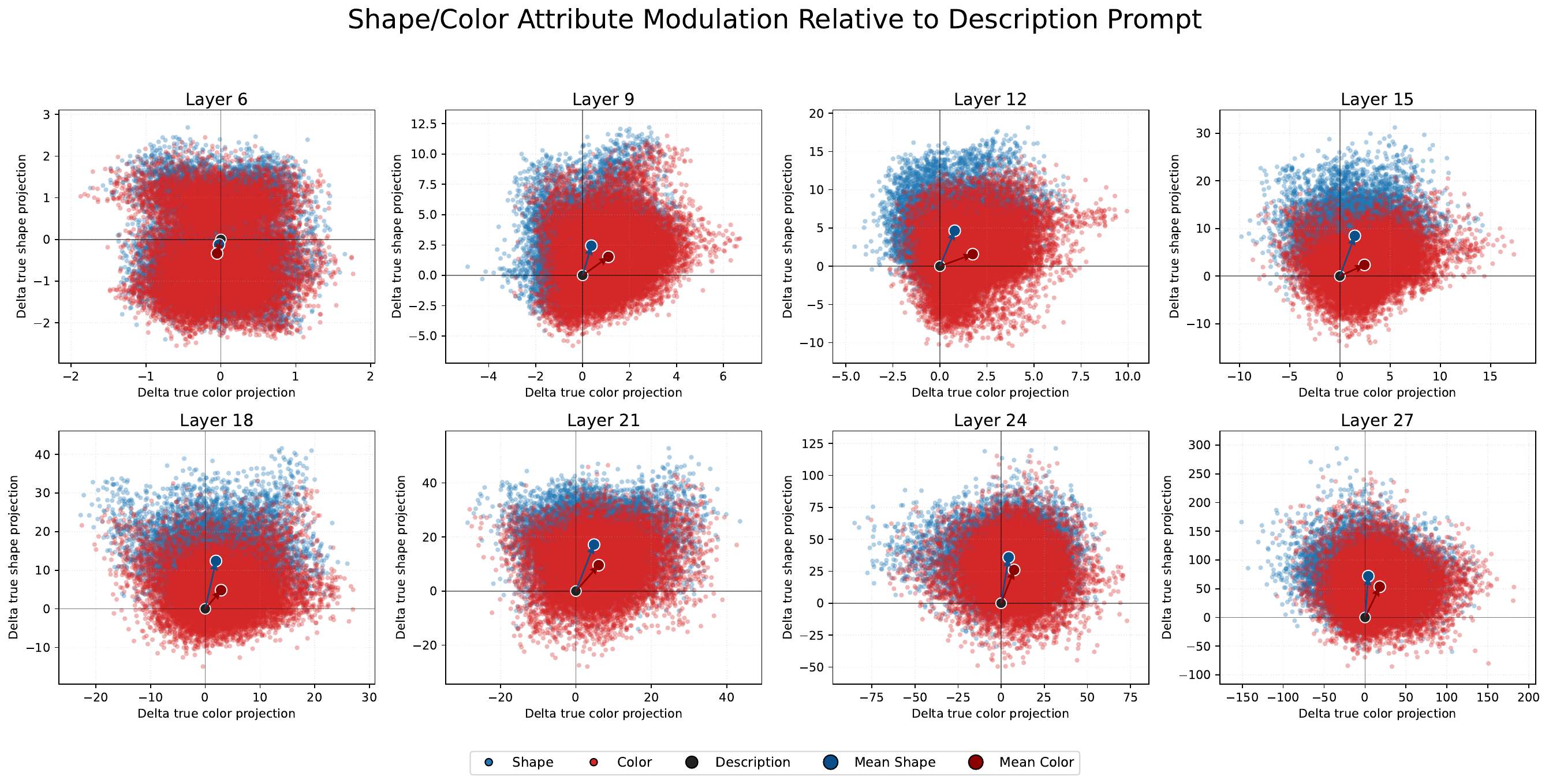}
    \caption{
Description prompt-relative shape/color displacement geometry across layers in Qwen. Each point corresponds to a single object. Blue points show shape-prompt displacements relative to the distractor prompt, while red points show color-prompt displacements. Large arrows denote the mean displacement for each prompt type. Across layers, compared to color prompts, shape prompts shift representations toward stronger true shape projections, while compared to shape prompts, color prompts shift representations toward stronger true color projections.
}
    \label{fig:qwen_scatterplot}
\end{figure*}

We show scatterplots across different layers of the model. Figure~\ref{fig:qwen_scatterplot} shows that for almost all layers, the mean shape has a higher true shape projection than the baseline and the mean color has a higher true color projection than the baseline. Additionally, the mean shape is to the upper left of the mean color, showing that the mean shape induces a stronger true shape projection and a weaker true color projection compared to the mean color.

\subsection{Modulation Plot}
\label{sec:shape_color_priming}
\begin{figure}[ht]
  \centering
  \includegraphics[width=\linewidth]{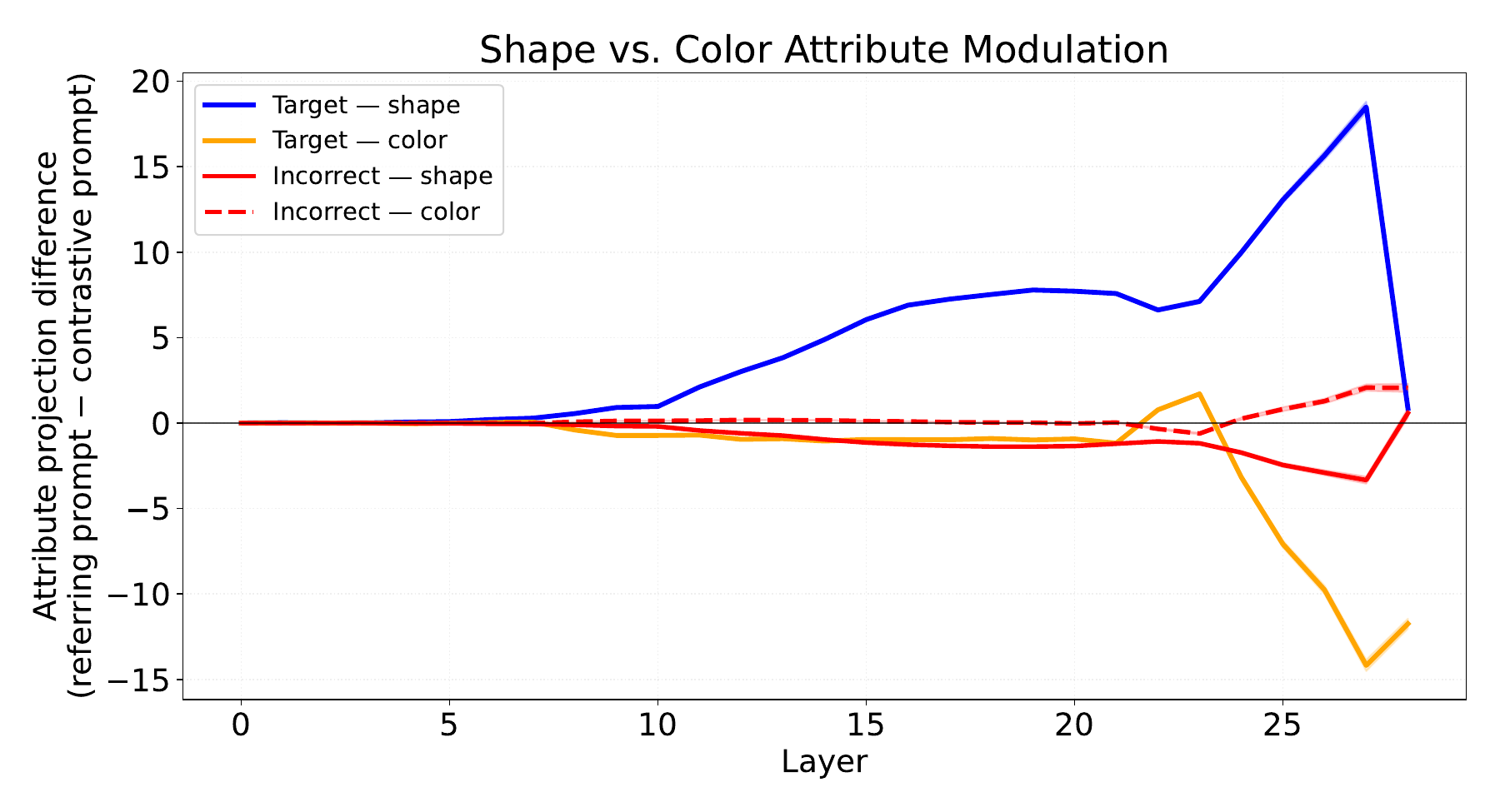}
  \caption{Explicit shape/color modulation on the standard synthetic dataset. Shape-focused prompts increase projections onto each object's true shape, while color-focused prompts increase projections onto each object's true color. Under the same contrast, absent shape projections become negative while absent color projections become positive, reflecting suppression of non-selected attribute representations. Absent-concept controls remain comparatively small throughout.}
  \label{fig:qwen_shape_color_priming}
\end{figure}

measuring modulation via
\begin{align}
\Delta_{\text{shape}}(o,l)
&=
\text{Proj}(o,s(o),p_{\text{shape}},l)
\nonumber \\
&\quad -
\text{Proj}(o,s(o),p_{\text{color}},l), \\
\Delta_{\text{color}}(o,l)
&=
\text{Proj}(o,c(o),p_{\text{color}},l)
\nonumber \\
&\quad -
\text{Proj}(o,c(o),p_{\text{shape}},l).
\end{align}
where $s(o)$ and $c(o)$ denote the true shape and color of object $o$. We additionally compute absent
shape/color controls by replacing $s(o)$ or $c(o)$ with a randomly chosen
attribute that does not appear anywhere in the image. 
The results are in  Figure~\ref{fig:qwen_shape_color_priming}. Under the
shape-focused versus color-focused contrast, true shape projections become
positive while true color projections become negative, indicating that the
model selectively strengthens representations of the queried attribute. The absent controls show the complementary pattern: absent shape projections become negative, while absent
color projections become positive. This flipped sign structure indicates active
suppression of non-selected attribute representations. Shape modulation is notably stronger than color modulation, suggesting that shape representations are more strongly reweighted by the prompt. 

\subsection{Representational similarity analysis}
\label{sec:rsa_appendix}

The shape vs.\ color modulation results in Section~\ref{sec:shape_color_priming} demonstrate 
that language queries amplify target attribute representations in object-level vision tokens by using their projection against concept vectors. 
Here we provide complementary evidence using Representational Similarity Analysis 
(RSA), examining how the geometry of vision token representations reorganizes as a 
result of the language query.

\paragraph{Setup.}
We construct a dataset of single object images by modifying the synthetic setup described in Section~\ref{sec:methods} to output one object instead of three. As before, each object is defined by a unique (color, shape) pair drawn from the same 6 colors and 6 shapes, and is patch-aligned to a grid cell. One image is generated per combination, yielding 36 images in total (6 colors $\times$ 6 shapes). For each image, the model is run under two prompts: a shape prompt ``What is the shape of this object?'' and a color prompt ``What is the color of this object?''. Vision tokens corresponding to the single object are extracted at each layer and mean pooled to produce one representation per image per layer per prompt, yielding two matrices of shape $(36, L, D)$. Pairwise cosine similarity matrices are then computed across all 36 images at each layer, separately under each prompt condition.

\begin{figure}[!t]
    \centering
    \includegraphics[width=\columnwidth]{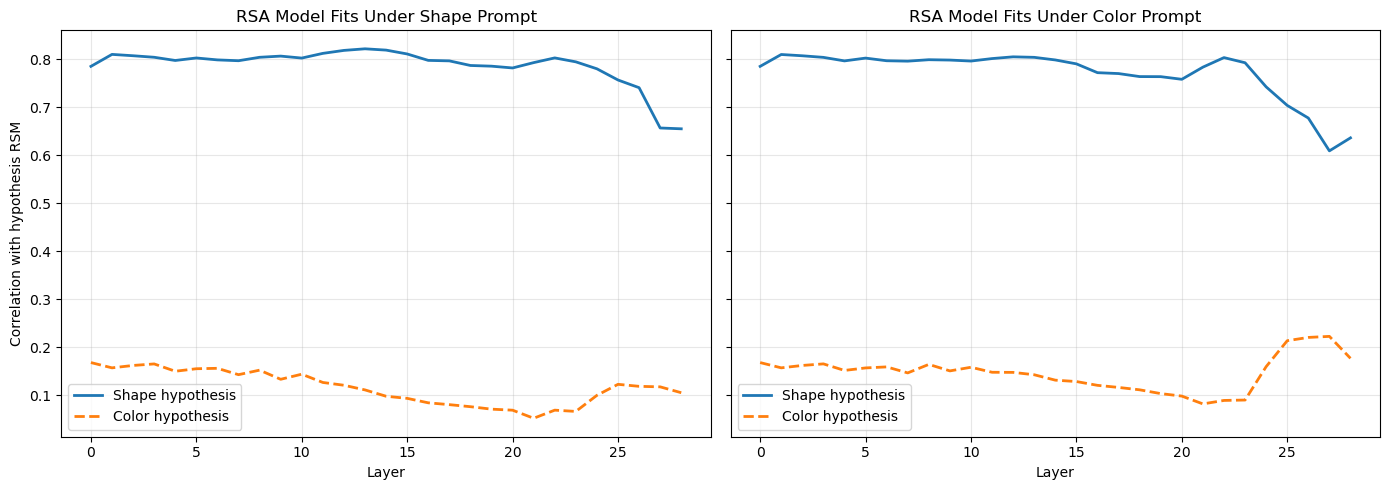}
    \caption{RSA model fit across layers. \textbf{Left:} Under the shape prompt, the shape hypothesis matrix correlates more strongly with the observed similarity structure than the color hypothesis matrix. \textbf{Right:} Under the color prompt, the shape hypothesis still correlates more strongly but its fit decreases, while the color hypothesis fit increases.}
    \label{fig:rsa_fit}
\end{figure}

\paragraph{RSA model fit.}
To quantify how well each attribute's structure explains the overall similarity geometry, we correlate each layer's similarity matrix with a binary hypothesis matrix $H$, where $H_{ij} = 1$ if images $i$ and $j$ share the same attribute value and $0$ otherwise. This is computed separately for shape and color. Figure~\ref{fig:rsa_fit} shows these correlation scores across layers under each prompt. Under both prompts, the shape hypothesis achieves a higher correlation than the color hypothesis. However, compared to the shape prompt, under the color prompt the shape hypothesis has a decreased fit while the color hypothesis has an increased fit. This confirms that the prompt modulates which attribute more strongly organizes the representational geometry of vision tokens, even when shape remains the dominant organizing principle across both conditions.

\begin{figure}[t]
    \centering
    \includegraphics[width=\columnwidth]{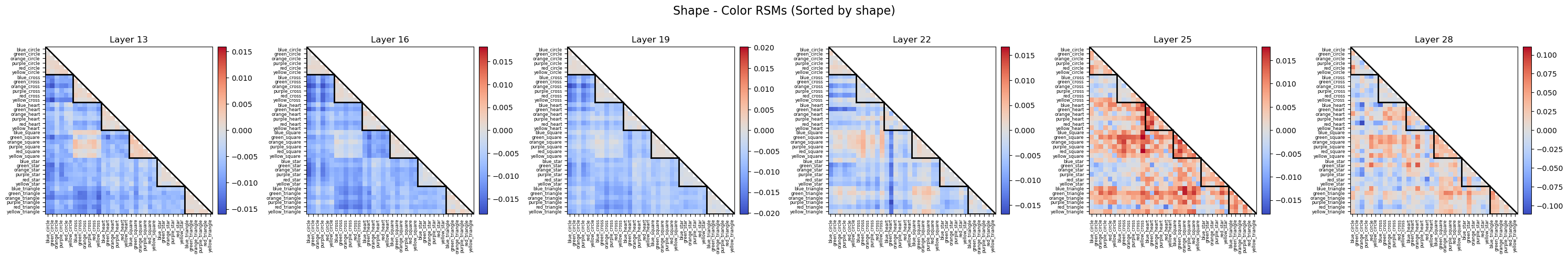}
    \vspace{0.5em}
    \includegraphics[width=\columnwidth]{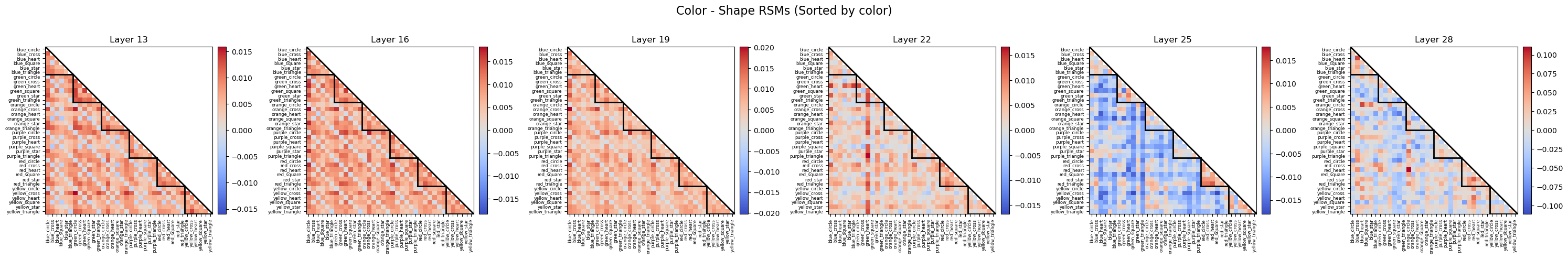}
    \caption{Difference RSMs across layers. Red indicates increased cosine similarity under the queried prompt relative to the contrasting prompt; blue indicates suppression. Black outlines mark same-attribute groups. \textbf{Top:} Shape Prompt $-$ Color Prompt sorted by shape. Same-shape blocks are neutral or red while off-diagonal regions are predominantly blue in layers 16, 19, and 22. \textbf{Bottom:} Color Prompt $-$ Shape Prompt sorted by color. The effect is subtler, but same-color blocks show red values by layers 25 and 28.}
    \label{fig:diff_rsms}
\end{figure}

\paragraph{Difference RSMs.}
To isolate query-induced changes in representational geometry, we compute difference RSMs: Shape Prompt $-$ Color Prompt when sorted by shape, and Color Prompt $-$ Shape Prompt when sorted by color. Red values indicate increased cosine similarity under the queried prompt relative to the contrasting prompt, blue indicates suppression, and black outlines mark same-attribute groups. As shown in Figure~\ref{fig:diff_rsms}, when sorted by shape, outlined same-shape blocks are consistently neutral or red while off-diagonal regions are predominantly blue in layers 16, 19, and 22, reflecting selective within-shape enhancement and cross-shape suppression. When sorted by color, the effect is subtler but by layers 25 and 28 the outlined same-color blocks show red values, indicating that color representations cluster more tightly under the color prompt than the shape prompt.

\section{Concept vector intervention}
\label{sec:concept_intervention}

The attribute amplification results in Section~\ref{sec:shape_color_priming}
show that querying for an attribute increases its projection onto the
corresponding concept vector at the referent object's position. We next test
whether this amplification is spatially localized to the referent object or
instead reflects a more global shift across vision tokens. To do so, we
intervene directly on object-level vision token representations by adding scaled
concept vectors at either the referent or non-referent object position and
measuring the downstream effect on answer confidence.

For each example, we measure confidence using the final-token logit gap between
the referent and non-referent object's correct attributes. For color, the gap is
the logit of the referent object's correct color minus the logit of the
non-referent object's correct color. For shape, the gap is defined analogously.
At layer $l$, we intervene on object $o$ by adding a scaled concept vector:
\begin{equation}
h_l(o,p)
\leftarrow
h_l(o,p)
+
\alpha V_l^a,
\end{equation}
where $V_l^a$ is the concept vector for attribute $a$ and $\alpha$ is the
intervention coefficient. We sweep over coefficients
$\alpha \in \{-120,-80,-40,0,40,80,120\}$ and measure the resulting change in
the final logit gap.

\begin{figure*}[t]
  \centering
  \begin{minipage}[t]{0.49\textwidth}
    \centering
    \includegraphics[width=\linewidth]{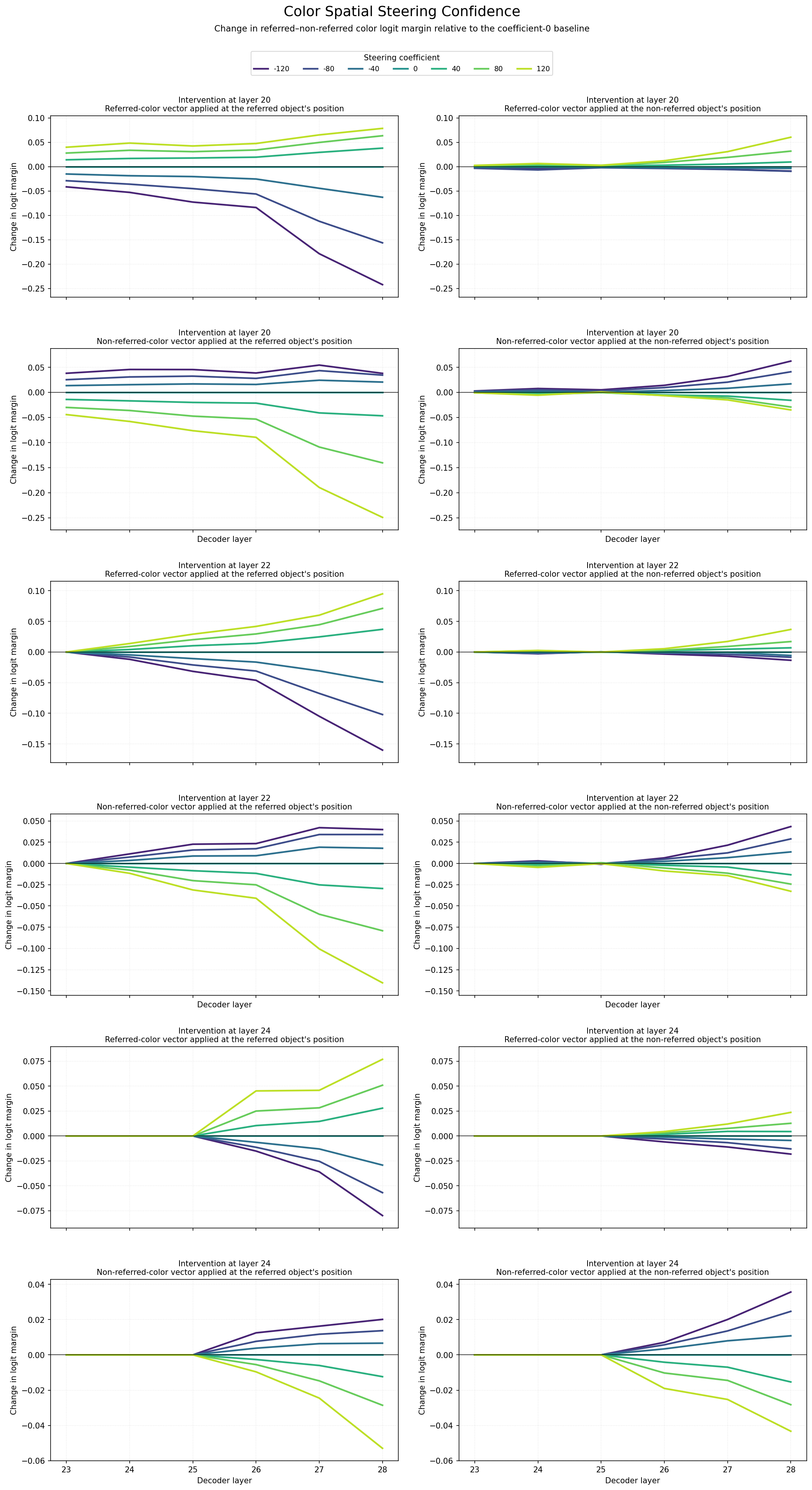}
    \small (a) Color
  \end{minipage}
  \hfill
  \begin{minipage}[t]{0.49\textwidth}
    \centering
    \includegraphics[width=\linewidth]{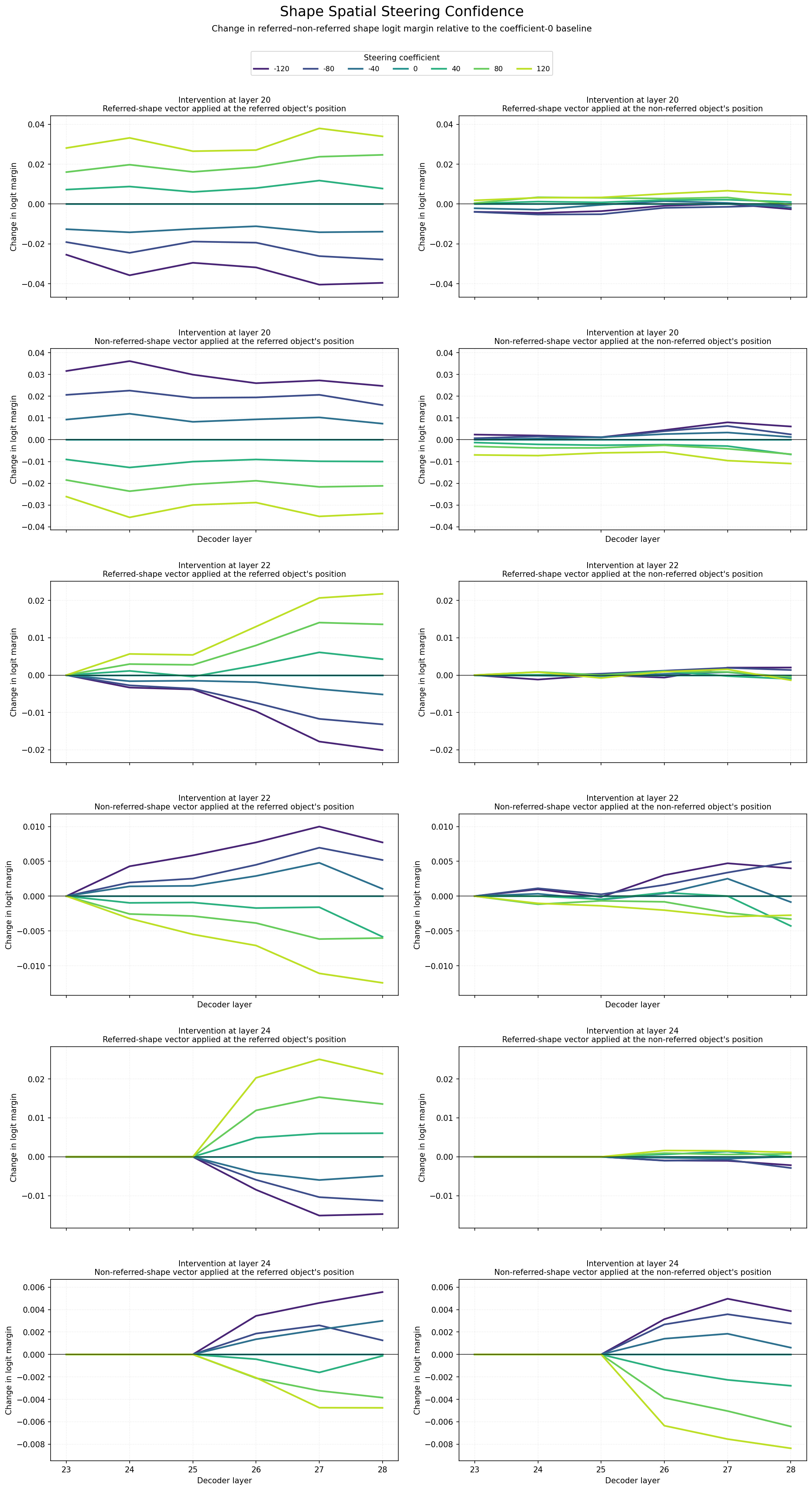}
    \small (b) Shape
  \end{minipage}

  \caption{Concept vector interventions at referent vs.\ non-referent object
  positions at layers 20, 22, and 24 for color (left) and shape (right). Intervening at the referent object
  position produces a substantially wider confidence steering range than
  intervening at the non-referent position, for both color and shape logit
  gaps. This confirms that attribute modulation is spatially localized to the
  referent object's vision tokens.}
  \label{fig:concept_intervention}
\end{figure*}

Results are shown in Figure~\ref{fig:concept_intervention}. Intervening at the
referent object position produces a substantially wider confidence steering
range than intervening at the non-referent position, for both color and shape
logit gaps. This effect is strongest in late layers, consistent with the
late-layer attribute amplification results in
Section~\ref{sec:referred_priming}. Additionally, adding the referred concept increases the logit gap while subtracting it decreases it. Together, these results confirm that attribute modulation is spatially localized: the referent object's vision tokens
are the specific site where query-induced attribute amplification is both
represented and causal.

\section{Shape Confidence from Freezing}
\label{sec:freeze_shape_gap}
\begin{figure}[H]
  \centering
  \includegraphics[width=\linewidth]{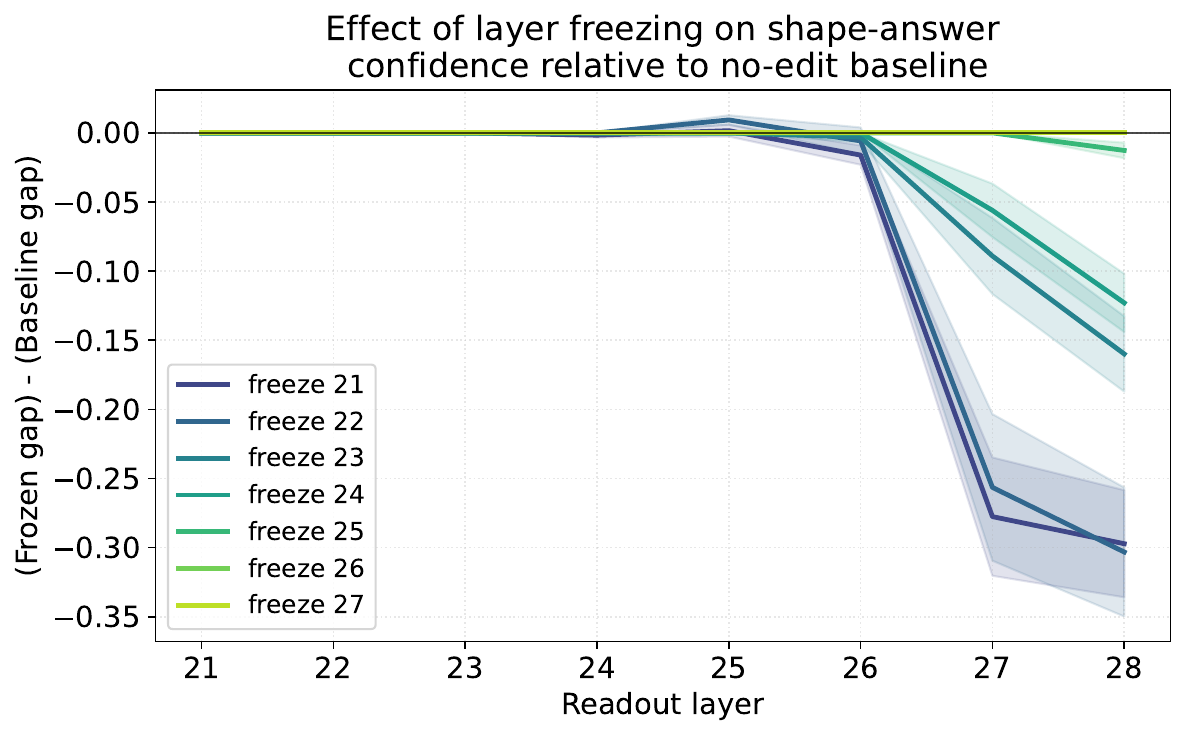}
  \caption{Freezing results for correct shape confidence}
  \label{fig:qwen_freeze_shape}
\end{figure}

We report the results for the effect of freezing at different layers on how much more the model is confident in the final shape compared to the baseline of no edit in Figure~\ref{fig:qwen_freeze_shape}. Similar to the color freezing results, freezing at later layers results in greater confidence compared to early layers, and freezing at any later layer results in less confidence than no freezing at all. 

\end{document}